\documentclass{article}
\usepackage[utf8]{inputenc}
\usepackage{url}
\usepackage{hyperref}

\usepackage{amsmath}
\usepackage{amssymb}

\usepackage{xcolor}
\usepackage{bm} 
\usepackage{bbm}
\usepackage{physics} 

\usepackage{subfig}  
\usepackage{graphicx} 

\usepackage{comment} 

\usepackage{tikz}
\usetikzlibrary{positioning}
\usetikzlibrary{calc}
\usetikzlibrary{shapes,arrows}
\usepackage{pgfplots}
\pgfplotsset{width=7cm,compat=1.15}
\usepgfplotslibrary{colorbrewer}

\usepackage{geometry}
\newgeometry{top=10mm, bottom=15mm}

\newtheorem{definition}{Definition}

\newtheorem{ex}{Example}

\usepackage{enumitem}

\newcommand{\dontshow}[1]{}

\newcommand{\ReLU}{{\rm ReLU}}

\usepackage{array}

\title{Equidistribution-based training of Free Knot Splines and ReLU Neural Networks}
\author{Simone Appella, Simon Arridge, Chris Budd, Teo Deveney, Lisa Maria Kreusser}
\date{\today}

\begin{document}

\maketitle

\begin{abstract}
We consider the problem of univariate nonlinear function approximation using shallow neural networks (NN) with a rectified linear unit (ReLU) activation function. We show that the $L_2$ based approximation problem is ill-conditioned and the behaviour of optimisation algorithms used in training these networks degrades rapidly as the width of the network increases. This can lead to significantly poorer approximation in practice than expected from the theoretical expressivity of the ReLU architecture and traditional methods such as univariate Free Knot Splines (FKS). Univariate shallow ReLU NNs and FKS span the same function space, and thus have the same theoretical expressivity. However, the FKS representation remains well-conditioned as the number of knots increases. We leverage the theory of optimal piecewise linear interpolants to improve the training procedure for ReLU NNs. Using the equidistribution principle, we propose a two-level procedure for training the FKS by first solving the nonlinear problem of finding the optimal knot locations of the interpolating FKS, and then determine the optimal weights and knots of the FKS by solving a nearly linear, well-conditioned problem. The training of the FKS gives insights into how we can train a ReLU NN effectively, with an equally accurate approximation. We combine the training of the ReLU NN with an equidistribution-based loss to find the breakpoints of the ReLU functions. This is then combined with preconditioning the ReLU NN approximation to find the scalings of the ReLU functions. This fast, well-conditioned and reliable method finds an accurate shallow ReLU NN approximation to a univariate target function. We test this method on a series of regular, singular, and rapidly varying target functions and obtain good results, realising the expressivity of the shallow ReLU network in all cases. We then extend our results to deeper networks.
\end{abstract}




\section{Introduction}

 \subsection{Overview}  Approximation of a target function $u({\mathbf x})$, ${\mathbf x} \in \Omega$ is one of the major applications of  {\em neural nets}  (NNs). Indeed,  a NN (regardless of its architecture) can  be regarded as a function generator for instance in the context of neural operators \cite{Kovachki2023}, and can be used  as means for solving both ordinary and partial differential equations \cite{karniadakis2021a}, \cite{E_DeepRitz}. In general, for an input  vector ${\mathbf x}$  to a NN,  the output of the NN is a nonlinear function evaluation $y({\mathbf x})$ where the corresponding function $y$ is defined on a suitable domain and has the same degree of smoothness as the NN activation functions. NNs  approximate the target function $u(\mathbf x)$ through a combination of linear operations and nonlinear/semilinear activation functions  which results in function approximations   nonlinear in their coefficients. The NN output $y({\mathbf x})$ is trained to approximate $u({\mathbf x})$. A key feature of NNs is that they are nonlinear approximators with provable expressivity, see  \cite{grohs_kutyniok_2022} and \cite{DeVore_Hanin_Petrova_2021} for instance for reviews on the  theoretical expressivity of deep NNs. By the Universal approximation theorem \cite{HORNIK1989}  any continuous function on a compact set can be approximated by a certain neural network, with a continuous activation function, to any accuracy. In principle a NN approximation also has the attractive feature that it can be 'self-adaptive' \cite{E_DeepRitz}, so that it may be able to account for singularities and rapid variations in the target function.
 A commonly used activation function is the   rectified linear unit (\ReLU), 
 defined as \ReLU$(x) = \max(x, 0) = (x)_+$. This is often used in practice due to its efficiency and simplicity. \ReLU \,NNs have a globally continuous and piecewise linear input-output relation. They are thus in one-dimension  formally equivalent to a piecewise linear free knot spline (FKS), with the \ReLU \,NN and FKS describing the same set of functions \cite{DeVore_Hanin_Petrova_2021}. 
 similar results hold in higher dimensions although the mapping one from one to another is rather more subtle. 
 For a \ReLU \,NN, the approximation error can  be explicitly linked to the depth $L$ and width $W$ of the network, with exponential rates of (error) convergence with increasing depth $L$ \cite{YAROTSKY2017}.

 \vspace{0.1in}
 
 \noindent However, as we will demonstrate, even with theoretical expressivity guarantees, this does not guarantee that the optimal solution is found by the training algorithm with the standard $L^2_2$ loss. This is because the resulting approximation problem is both  badly conditioned and highly non-convex. 
 As a result when training we may either not see convergence, or convergence to a sub-optimal solution, with errors which are many orders of magnitude worse than they should be. This is either because the optimal approximation with high expressivity is not found, or the training process is far too slow and is terminated too early. Similar behaviour is seen in PINNS used to approximate the solution of differential equations, see for example \cite{CBSFEM}.  In this paper we will draw on results from classical adaptive approximation theory to show how the training approach for such a NN can be improved significantly, both by using a more appropriate loss function to overcome the convexity issues (so that we can find the most expressive approximation), and also by preconditioning the NN representation to overcome the ill-conditioning of the training procedure. 
 By doing so we can achieve the theoretical expressivity of the NN in cases where the usual training method fails to deliver a good approximation. 
 

 \vspace{0.1in}

\subsubsection{Classical approximation theory}
 
 Classical approximation theory has long considered the  problem of efficiently approximating a target function $u({\mathbf x}), {\mathbf x} \in \Omega$ \cite{devore1993constructive,Powell_1981}. Examples include the use of combinations of basis splines with localised support, for function interpolation or least squares approximation, such as the Galerkin methods used in the Finite Element Method. Such classical function approximations are usually {\em linear} in their coefficients, as the weights ${\mathbf w}$  of a given set of basis functions are optimised and the function approximation lies in a linear function space. Typically the basis functions themselves are piece-wise polynomials (splines), defined over a {\em fixed} mesh (in d-dimensions), or more simply a set of {\em knots} ${\mathbf k}$ in one-dimension, where the form of the local polynomial approximation changes. These methods have provable accuracy \cite{DeVore_1998} and there exist efficient, and well-conditioned, algorithms for calculating the optimal weights that reliably converge to a unique solution, often either solving well-conditioned linear systems of equations (directly or by iteration), or by a simple quadratic minimisation problem. However, such linear methods often lack accuracy, and may require a large number of coefficients to achieve a good level of approximation for a function with complex, or singular behaviour. An important example of such an approximation in one dimension, is the piecewise linear interpolant $\Pi_1 u$  defined over a {\em uniform mesh} with $N$ points. If $u \in H^2(\Omega)$ we then have, as $N \to \infty$
 \begin{equation}
 N^2 \|\Pi_1 u - u\|_{2} \le C  \|u\|_{H^2(\Omega)}.
 \label{sept1}
 \end{equation}
 If $u$ is not in $H^2(\Omega)$ this estimate fails and we see a sub optimal rate of convergence
 Even if $u \in H^2(\Omega)$ when approximating a rapidly varying function good convergence is only seen when the local knot spacing is smaller than the smallest length scale of the target function,

 \vspace{0.1in}
 
 \subsubsection{Free Knot Splines (FKS)} Significantly greater accuracy, with far fewer numbers of coefficients, is achieved by using an (adaptive) nonlinear approximation method \cite{deboor} such as  a Free Knot Spline (FKS) in one-dimension,  or adaptive mesh methods \cite{HuangRussel}, \cite{M2M} in higher dimensions. In such approximations both the weights ${\mathbf w}$ and the knot locations ${\mathbf k}$/mesh are optimised. 
Even though such methods are  nonlinear in the coefficients, effective a-priori and  a-posteriori error estimates and interpolation error estimates have been established linked to the regularity of the meshes over which the approximations are defined \cite{HuangRussel}. A key advantage of using such an adaptive approximation over a standard spline approximation, is that for a fixed number $N$ of knots/mesh-points ${\mathbf k}$ a much higher accuracy can be achieved by exploiting the extra degrees of freedom given by moving the points ${\mathbf k}$ \cite{DeVore_1998}, \cite{HuangRussel}, \cite{M2M}, which is of importance when approximating functions with rapid variation and/or singularities.  A special case are linear interpolating FKS (IFKS) and similar interpolating approximations in higher dimensions. In these approximations the weights are evaluations of the target function $u(k_i)$ at the knot/mesh-point locations, and as such the optimisation reduces to finding knot/mesh-point locations ${\mathbf k}$ only. 
 Tight convergence results in various norms have been developed in a general dimension $d$ for the best interpolating piecewise linear approximation $\Pi_1 u$ to a general target function $u$ as $N \to \infty$. Summaries are given in \cite{HuangRussel}, \cite{DeVore_1998} and later in this paper.  In particular, in one-dimension, the optimal approximation error between a general target function $u$ (even one of limited smoothness)  and a linear interpolating FKS  $\Pi_1 u$  with an {\em optimal choice of $N$ knots} satisfies 
 \begin{equation}
 N^2 \|\Pi_1 u - u\|_{2} \le  C \left(\int_{\Omega} (u'')^{2/5} \; dx \right)^{5/2}.
 \label{sept2}
 \end{equation}
 The estimate (\ref{sept2}) is in general much better than that in (\ref{sept1}) and applies for functions which are not in $H^2(\Omega)$.
Whilst the 
IFKS is, in general, far from the best approximation, the optimal knots for the best IFKS are usually close to those for the best FKS.
 Training a general FKS approximation then becomes a {\em two stage process}. The first being the (nonlinear) problem of finding the optimal knot/mesh points and the second the (usually) well conditioned problem of finding the correct weights.  We show in this paper that this combination gives excellent approximations with high accuracy. 

\vspace{0.1in}

\subsubsection{Combining NN and classical adaptive approximation} The complementary advantages of FKS and \ReLU \,NNs, and their formal equivalence, motivates us to exploit the relationship between shallow \ReLU \,NNs and splines, and in particular both the direct correspondence between the knot points of the FKS and the breakpoints (where the derivative changes discontinuously) of the shallow \ReLU \,NN, and also between the coefficients of the FKS and the \ReLU \,NN. Using the spline lens, shallow univariate \ReLU \,NNs have been investigated in terms of their initialization, loss surface, Hessian and gradient flow dynamics \cite{Sahs2022}.  A \ReLU \,NN representation for continuous piecewise linear functions has been studied in \cite{He2020}, with a focus on continuous piecewise linear functions from the finite element method with at least two spatial dimensions. A theoretical study on the expressive power of  NNs with a \ReLU~activation function in comparison to linear spline-type  methods is provided in \cite{ECKLE2019232}. The connection between NNs and splines have also been considered in other contexts, for instance when learning activation functions in deep spline NNs \cite{Bohra2020}.

\begin{figure}[htbp]
  \centering
   \includegraphics[width=0.4\textwidth]{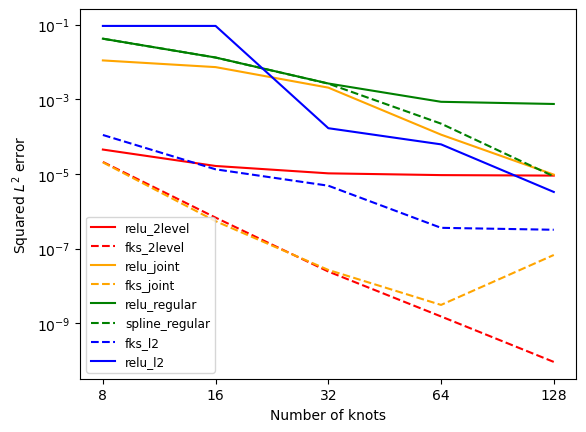}
      \includegraphics[width=0.45\textwidth]{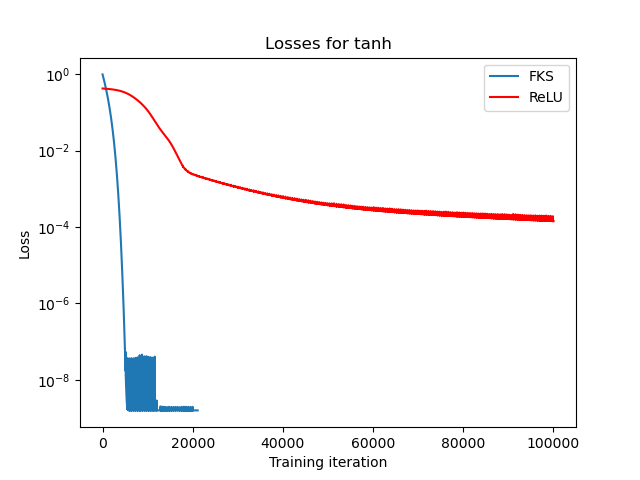}
   \caption{(Left) A comparison of the convergence of the approximation of the target function $u(x) = \tanh(100(x-1/4))$  with various architectures. 
   This shows (green) the poor performance of the \ReLU \,NN trained using the regular $L_2^2$ loss function, the better, but still poor, performance of the regular spline with the same loss function (green dashed).  We see the still poor performance of the ReLU NN and the FKS trained using the standard $L_2^2$ loss function (blue,blue dashed) We also see the far better performance of the FKS trained with a two level approach (red dashed) but the much poorer performance of the ReLU NN using the two-level approach without pre-conditioning (red). A preconditioned \ReLU \,NN architecture trained in the two-level approach shows similar optimal performance to the FKS. (Right) A comparison of the training times and convergence  of the weights of the FKS and a shallow \ReLU \,NN for the target function $u(x) = \tanh(100(x-1/4))$ with $N=64$, following the calculation of the knots of the interpolating FKS. The FKS trains very rapidly, whereas the \ReLU \,NN trains slowly due to ill-conditioning.}
    \label{fig:cfeb1ff}
\end{figure}

To motivate this comparison we consider the approximation of the target function 
$$u(x) = \tanh(100(x-0.25)), \quad x \in [0,1].$$
This function has a narrow boundary layer at $x = 1/4$ which must be resolved to give a good approximation. We consider various methods of approximating this function, in particular (i) 'regular' spline and ReLU approximations with fixed uniform knots/breakpoints (ii) a shallow \ReLU \,NN and an FKS trained with the 'standard' $L_2$ loss function (iii) a two-level approach with the \ReLU \,NN trained to find the optimal break points and then the optimal weights (without preconditioning) and similarly a FKS trained to first find the optimal knots and then the optimal weights.  In Figure \ref{fig:cfeb1ff} we plot (left) the resulting $L_2^2$ loss of the trained networks as a function of the number of knots/width of the network. We can see from the this figure that the \ReLU and regular spline perform poorly, as does the ReLU NN trained on the $L_2$ loss function. The FKS trained on the $L_2$ loss function performs better but still sub-optimally with poor convergence for large $N$. The FKS trained using the two-level approach performs essentially optimally with rapid convergence, but the same approach for the ReLU NN gives a much worse result if no preconditioning is used. The difference in approximation error between the \ReLU \,NN approximation and the correctly trained FKS is striking with a difference in error of many orders of magnitude. Similarly the difference between the regular spline error and the FKS error is also striking due to the incorrect knot location. This latter problem is well conditioned. To see the role that ill-conditioning of the \ReLU\,network plays we consider
using both a FKS with $N=64$ knots and the formally equivalent shallow \ReLU \,NN of width $W = 64$ which has $N$ break points. 
In Figure \ref{fig:cfeb1ff} (right) we compare the training, using the $L_2^2$ error, of both in the case where the knots/break points are fixed to the optimal values for the IFKS. 
It is clear from Figure~\ref{fig:cfeb1ff} that with this set of knots the FKS trains very much more quickly, and to a much more accurate approximation than the \ReLU \,NN for this target function. This, is due to the poor conditioning of the \ReLU \,NN approximation problem for large $N$ associated with the condition number of the normal equations of the training process.  Only by finding the breakpoints of the \ReLU\,network first and then preconditioning the problem of finding the weights can the same accuracy as the FKS be realised.

\subsection{Contributions}

\noindent We  analyse the problem of function approximation from the perspective of training both a \ReLU \,NN and a FKS/adaptive spline, and hence improve the conditioning, training and accuracy of the \ReLU \,NN architecture 

\vspace{0.1in}

\noindent Our initial theoretical analysis considers the most simple case of univariate approximation using a shallow \ReLU \,NN and an FKS. 
Whilst this is removed from the usual practice of using deep networks on higher dimensional problems, the simplicity of this problem allows us to conduct a careful mathematical analysis, which gives much insight into the numerical results that we obtain for more general (deep and higher dimensional) problems.

\vspace{0.1in}

\noindent We  show rigorously that in this special (univariate, shallow) case the \ReLU \,NN approximation problem is poorly conditioned as the network width $W$ increases. As a result the training process is very slow, leading either to a poor approximation of the target function, or simply to no convergence. We  propose a novel two-level approach to the nonlinear problem of training (using the Adam optimiser) the univariate FKS approximation with weights ${\mathbf w}$ and (free) knots ${\mathbf k}$). This problem is  tackled by constructing an effective loss function based on the equidistribution principle for the knot locations of the interpolating FKS, and minimising this using the Adam optimiser. We then solve the provably well conditioned problem of finding the weights and (close-by) knots of the FKS. 
We then show  that this approach can be used to train the shallow  \ReLU \,NN, firstly by finding breakpoints of the \ReLU \,NN (as the knots of the interpolating FKS), and then the scaling factors of the \ReLU~functions. We show that this latter problem can be made well conditioned for large $W$ by a linear preconditioning of the \ReLU \,NN, transforming it into an FKS, and then training the resulting transformed system. 
Using a variety of (regular, singular and rapidly varying) target functions, we demonstrate numerically that this two-stage training procedure means that the shallow univariate \ReLU \,NN can be trained reliably to give a highly accurate approximation, achieving the theoretical expressivity of these networks. 

\vspace{0.1in}

\noindent We then extend our results to univariate approximations using 
more general activation functions, and also to deeper networks. The complexity of these systems makes an analytical investigation difficult. However we show, by numerical experiments, that the training of the \ReLU \,NN is still ill-conditioned and (if it converges at all) gives a sub-optimal solution. In contrast methods based on spline approximation can be derived which have high accuracy.


\subsection{Outline} The remainder of this paper is structured as follows. In Section \ref{sec:Notation} we  describe  univariate \ReLU \,NNs, as well as free knot splines (FKS)  as nonlinear function approximators. In Section 3  we show the formal equivalence of shallow \ReLU \,NN and the FKS representations and the links between the breakpoints/knots and the parameters/weights of these respective architectures. Section \ref{sec:loss_training} defines the loss functions often used based on the mean-squared error and discusses the difficulty of optimising the approximation in using these functions to train the knots/breakpoints and weights together. In Section 5 we focus on the problem of finding the optimal knots for an interpolating FKS and introduce a loss function based on equidistribution as a way of guiding the training. In Section 6 we consider the problem of finding the optimum weights having found the knots (as in Section 5). We show that this problem is well-conditioned for the FKS representation of the solution, but is ill-conditioned for the shallow \ReLU \,NN architecture. In Section 7 we then introduce the idea of two-level training for the knots and then the weights of an FKS, and then show how, after preconditioning, it leads to an effective optimisation approach for the shallow \ReLU \,NN.
 Numerical results for the one dimensional approximation using both a FKS and a shallow network for a range of target functions are shown in Section \ref{sec:Numerical Results} where we investigate the impact of the choice of loss function, the quality of the function approximation, the evolution of the knot locations as well as the conditioning and the convergence of the loss training. Extensions of this work to different activation functions, deep networks and to higher dimensions are discussed in Section 9.
 Finally, in Section \ref{sec:conclusion} we will draw our conclusions based on the results, and suggest areas of further investigation.

\section{ReLU \,NN and free knot splines (FKS) as nonlinear function approximators}
\label{sec:Notation}

In this section, we define univariate \ReLU~Neural Networks (NN) and free knot splines (FKS). 

\subsection{\ReLU~Neural Networks (\ReLU\,NN)} 

 \noindent We start by defining  a  standard feed forward neural network with a one-dimensional input $x_0\in \mathbb R$ and one-dimensional output $y(x_0,\mathbf{\theta})$ for some vector of parameters $\mathbf{\theta}$, described further below. We assume that this network has a width $W$ and a depth $L$ and uses a rectified linear unit (\ReLU)  $\sigma(x) = (x)_+$  as  activation function which acts element wise on  the input vector.
Defining the network input by $x_0\in\mathbb{R}$, the 
first hidden layer is defined by
$\boldsymbol x_{1} = \sigma \left(\boldsymbol {A_0} x_0 +\boldsymbol b_0 \right) \in \mathbb{R}^W$
for $\boldsymbol A_0,\boldsymbol b_0\in \mathbb R^W$ and the
NN with $L$ hidden layers is then iteratively defined by
\begin{equation}
\boldsymbol x_{k+1} = \sigma \left(A_k \boldsymbol x_k + \boldsymbol b_k \right) \in \mathbb{R}^W 
\label{eq:Shallow-net}
\end{equation} 
for $k\in\{ 1, \ldots, L-1\}$,
with $A_k \in \mathbb{R}^{W \times W}$ and $\boldsymbol b_k \in \mathbb{R}^W$. The NN output is given by
\begin{equation}
    y(x,{\mathbf{\theta}}) = \boldsymbol A_L \boldsymbol x_L +  b_L \in \mathbb R,
    \label{eq:Shallaw-net3}
\end{equation}
with $\boldsymbol A_L \in \mathbb{R}^{1\times W}$ and $b_L \in \mathbb{R}$.
In the simplest case we can consider a {\em shallow} \ReLU \, network with width $W$ and depth $L=1$, otherwise the network is {\em deep}. The case $L=1$ is closest to standard approximation, but the  power and expressivity of the NN often arises when $L > 1$, although  such networks are harder to train and to analyse. In this representation we define the set of parameters ${\mathbf{\theta}}  = \{A_k,\boldsymbol b_k\colon~ k=0,\ldots,L\}$.
For a given range of parameters ${\mathbf{\theta}}$ we define $\Gamma_{W,L}$ to be the set of all functions $y(x)$ which can be represented by the NN. Observe that $\Gamma_{W,L}$ is {\em not} a linear vector space and that $A_k$ and $\boldsymbol b_k$ are not sparse in general.
We apply this NN to approximate a function $u(x) \colon  [0,1]\to \mathbb R$ by finding an appropriate choice of $\mathbf{\theta}$. We assume that there exists a batch of training data $\{y_{x_i}\}_{i=0}^{s-1}\subset \mathbb{R}^s$ at training points $\{x_i\}_{i=0}^{s-1} \subset [0,1]$ of size $s$. Typically this uses the standard means of training a NN to do this, including the use of a squared-error $L_2$ loss function given by $L = \|y - u\|_2^2$ and an optimiser such as Adam \cite{KingBa15}. However, as we will see, crucial to the effectiveness of the training procedure is the use of a good initialisation to start the approximation, an effective (convex) loss function, and a careful preconditioning of the problem. We shall demonstrate this using the examples in this paper, and will also show how an effective  two-level training method  can be constructed, using a suitably convex loss function combined with preconditioning.

\subsection{Linear spline approximations}\label{sec:splines}

 Let $N\in \mathbb N$ be given and let
$0 = k_0 < k_1 < \ldots k_i < \ldots < k_{N-1} = 1$
be a set of $N$ {\em knots} in the interval $[0,1]$. For $i=0,\ldots, N-1$, we define a {\em linear spline} $\phi_i(x)$ to be the unique piecewise linear function so that
$\phi_i(k_i) = 1$, and $\phi_i(k_j)= 0$ if $i \ne j.$
Each spline $\phi_i$ is weighted by some weight $w_i$, and we denote by ${\mathbf{\psi}_{\rm FKS}} = \{ ( w_i, k_j)\colon i=0,\ldots, N-1,~ j=1,\ldots,N-2 \}$  the set of parameter values. Note that we exclude $k_0$ and $k_{N-1}$ in the definition of $\mathbf{\psi}_{\rm FKS}$ as they are set as 0 and 1, respectively.
\begin{definition}[Free knot splines (FKS)]
A  \emph{free knot spline (FKS)} is a piecewise linear approximation to the function $u(x)$ defined over the set of linear splines  and is  given by
\begin{equation}
    y(x,\mathbf{\psi}_{\rm FKS})= \sum_{i=0}^{N-1} w_i \phi_i(x), 
    \label{c2}
\end{equation}
where we assume that we are free to choose both  the knots $k_i, ~i=1,\ldots,N-2$, in the linear splines $\phi_i$ and the weights $w_i, i = 0 \ldots N-1$.
\end{definition}
\noindent We denote by $\Sigma_{N,1}$  the set of all piecewise linear functions of the form \eqref{c2}. Observe that the FKS has $2N-2$ degrees of freedom ($N-2$ for the knots and $N$ for the weights). Like the NN space $\Gamma_{W,L}$, the approximation space $\Sigma_{N,1}$ is nonlinear. 
 Similarly to the NN we can train a FKS to approximate a function. This is a nonlinear, non-convex problem to find the full set of weights ${\mathbf w}_i$  
 and knots ${\mathbf k}_i$.

\vspace{0.1in}

\noindent  In the more usual case of piecewise linear approximation, the values of the knots $k_i$ are  {\em fixed} and we refer to such approximations as either \emph{fixed knot piecewise linear approximations} or \emph{regular approximations}. The best function approximation is given by the well-conditioned problem of finding the optimal choice of {\em weights} $w_i$ for $i=0,\ldots,N-1$ and we denote by $\mathbf{\psi}_{\rm PWL}=\{ w_i\colon i=0,\ldots,N-1\}$ the associated parameter values. 
\begin{definition}[Fixed knot piecewise linear interpolant (PWL)]
Such a \emph{fixed knot piecewise linear approximation} is a piecewise linear approximation to the function $u(x)$ defined over the set of linear splines  and is  given by
\begin{align}\label{eq:pwl}
y(x,\mathbf{\psi}_{\rm PWL})= \sum_{i=0}^{N-1} w_i    \phi_i(x)
\end{align}
where  we assume that we are free to choose  the  coefficients $w_i$, but   the knots $k_i, ~i=1,\ldots,N-2$, in the linear splines $\phi_i$ are assumed to be fixed.
\end{definition}
\noindent Note that \eqref{eq:pwl}
is a linear function of the coefficients $w_i$ and the set of all such possible functions is a linear vector space. The values of $w_i$ can  be found simply by minimising the least squares loss. Equivalently, a linear matrix problem of the form $\boldsymbol P \boldsymbol w = \boldsymbol q$ for the coefficients $\boldsymbol w=(w_i)_{i=0}^{N-1}\in \mathbb R^N$ can be solved, where the (well conditioned) sparse matrix $\boldsymbol P\in \mathbb R^{n\times n}$  and vector $\boldsymbol q\in \mathbb R^n$ are given. Whilst this is a simple process and widely used, for example in the Finite Element Method,  such approximations lack expressivity, especially if the target function $u(x)$ is singular, or has natural length scales which are smaller than the smallest separation between the knots. The additional degrees of freedom given by choosing the knots $k_i$ in the FKS  leads to a very significant increase in the accuracy of the approximation. 

\vspace{0.1in} 

\noindent  A distinguished subset of the set of free knot splines are the {\em interpolating free knot splines} where the set of knots $k_i$ for $i = 1, \ldots, N-2$ can be freely chosen, but the weights are given by $w_i = u(k_i)$  We denote the associated parameter values by $\mathbf{\psi}_{\rm IFKS}=\{ k_i\colon i=1,\ldots,N-2\}$.
\begin{definition}[Interpolating free knot splines (FKS)]
The \emph{interpolating free knot splines (FKS)} $y(x,\mathbf{\psi}_{\rm IFKS}) = \Pi_1 u(x)$ for the target function $u(x)$ is given by
\begin{equation}
y(x,\mathbf{\psi}_{\rm IFKS}) \equiv \Pi_1 u(x) = \sum_{i=0}^{N-1}  u(k_i) \phi_i(x),
\label{oct1}
\end{equation}    
where we assume that we are free to choose   the knots $k_i, ~i=1,\ldots,N-2$, in the linear splines $\phi_i$.
\end{definition}

 \noindent The interpolating FKS is easier to construct than the best FKS approximation for $u(x)$ as the coefficients $w_i$ of the FKS are given directly by $w_i = u(k_i)$ and only the knots need to be determined. It is thus not as expressive as the best FKS, but has  the same rate of convergence as $N \to \infty$ for classes of singular functions \cite{HuangRussel}, and it is a much better approximation in general (especially for smaller values of $N$) than a fixed knot piecewise linear approximation. Observe that  finding the interpolating FKS  still requires solving a nonlinear problem for the knots. We will make extensive use of this function as a way of guiding the approximation procedure and also assessing the accuracy of a more general FKS/\ReLU \,NN approximation to $u$. In particular $\Pi_1 y$ is a very good function to take as the initial start of the optimisation procedure for finding either the best \ReLU \,NN approximation or the best FKS approximation to $u$. This is because the {\em optimal} set $K$ of knot points is constructed for $\Pi_1 u$ appears from extensive calculations to be very close to the optimal set of knot points for the FKS, and equivalently the breakpoints of the \ReLU \,NN representations.

\section{Relations between the \ReLU \,NN and the FKS representations}\label{sec:comparison_ReLUFKS}

 Any \ReLU \,NN  gives a function $y(x)$ which is piecewise linear, and is smooth (linear) everywhere apart from at a set of  breakpoints   where it changes gradient and has undefined curvature. These points are equivalent to the   knot points  $k_i$ of the FKS if they lie in the range of values of $x \in [0,1]$  over which the target function $u(x)$ is defined. In this sense the \ReLU \,NN and the FKS are formally equivalent. For a deep NN the map between ${\mathbf{\theta}}$ and $\{k_i\colon i=0,\ldots,N-1\}$ is complex and is related to the solution of a system of algebraic equations \cite{DeVore_Hanin_Petrova_2021}. More precisely, the authors show that a NN of width $W$ and depth $L$  is formally equivalent to a FKS of size $N$ (where $N$ is the number of breakpoints) for $N <  (W+1)^L$, and hence they can have the same expressivity, even though this is rarely seen in practice. Due to the linear dependencies between the linear pieces,  the NN  space  $\Gamma_{W,L}$ is usually much smaller than the FKS space $\Sigma_{N,1}$ with $N \ll  (W+1)^L$. 
 Accordingly we focus on the case of shallow NN where the correspondence is direct. 

\subsection{Equivalence of shallow ReLU \,NN and the FKS representations}

 Both a ReLU \,NN and an FKS are piecewise linear functions of $x$ and we can use the methodology of training one to inform the training of the other.  For a shallow network with $L=1$ and $N=W$, we have breakpoints in the \ReLU representation, which are equivalent to knots in the FKS at 
\begin{equation} 
k_i = -\frac{(\boldsymbol b^0)_i}{(\boldsymbol A_0)_i},\quad i=0,\ldots,W-1 \quad \mbox{if} \quad k_i \in [0,1],
\label{c1}
\end{equation}
where $(\boldsymbol b^0)_i$ and $(\boldsymbol A_0)_i$ denote the $i$th components of vectors $\boldsymbol b_0$ and $\boldsymbol A_0$.
For any ordered set of knots $k_i$, with $0 = k_0 < k_1 < \ldots < k_{N-2} < k_{N-1} = 1$ we can represent the piece-wise linear basis functions $\phi_i(x)$ of the FKS using the formula presented in  \cite{DeVore_Hanin_Petrova_2021} so that 
\begin{equation}
\phi_i(x) = \alpha_i \ReLU(x - k_{i-1}) 
-\beta_i \ReLU(x - k_i) + \gamma_i \ReLU(x - k_{i+1})
\label{dv1}
\end{equation}
for $i=1,\ldots,N-2$,
where 
\begin{equation}
    \alpha_i = \frac{1}{k_i - k_{i-1}}, \quad  \beta_i = \frac{k_{i+1} - k_{i-1}}{(k_{i+1} - k_i)(k_i - k_{i-1}) }, \quad \gamma_i= \frac{1}{k_{i+1} - k_i}.
    \label{cjan13c}
    \end{equation}
    Similarly,
    $\phi_0(x) = \ReLU(k_1-x)/k_1, \quad \phi_{N-1}(x) = \ReLU(x - k_{N-2})/(1-k_{N-2}).$
This implies that FKS representation in terms of the basis functions can also be given in terms of \ReLU~functions as follows
\begin{equation}
y(x,\mathbf{\psi}_{\rm FKS}) = \sum_{i=0}^{N-1} w_i \phi_i(x)  \equiv w_0 \ReLU(k_1-x)/k_1 + \sum_{i=0}^{N-2} c_i \ReLU(x - k_i)
\label{cnov123}
\end{equation}
where the \ReLU coefficients $c_i$ are rational functions of the knots $k_i$ and weights $w_i$ of the FKS. Taking careful note of the representation of the basis functions at the two ends of the interval we obtain:
\begin{equation}
\begin{split}
c_0 &= \alpha_1 w_1,\quad
    c_1 = \gamma_1 w_2 - \beta_1 w_1, \\
    c_i &= \gamma_{i} w_{i+1} - \beta_i w_i + \alpha_{i} w_{i-1}, \quad i = 2,\ldots,N-2.
    \label{cjan13d}
    \end{split}
\end{equation}
 Hence the two representations of the functions by the \ReLU\,NN and by FKS are formally equivalent, with the equivalence of the \ReLU breakpoints with knot points, and with the {\em tri-diagonal linear} map (\ref{cjan13d}) between the weights ${\mathbf w}$ and scaling constants ${\mathbf c}$. Hence (in principle) both networks are equally {\em expressive}. However, the information from the respective parameter sets ${\mathbf{\theta}} $ and ${\mathbf{\psi}_{\rm FKS}}$ is encoded differently,  and is critical for their training and in particular for the well-posedness of the solution representation and the associated training algorithm.  In general, the coefficients $w_i$ of the FKS are ${\mathcal O}(1)$, and may be monotone and of constant sign. However, if,  for example,  $k_i - k_{i-1}$ is small (which is often the case if we wish to approximate a function with a high gradient or high curvature) then the coefficients in (\ref{cjan13c}) may well be  large, and of non-constant sign, even if $u(x)$ has constant sign. As a consequence, the \ReLU \,NN representation of a function may  be ill-conditioned, even if the FKS representation is well-conditioned. 

\subsection{Weights for the IFKS and equivalent ReLU \,NN}
 As an example, we consider the interpolating FKS $\Pi_1 u$ on the set of knot points for which $w_i = u(k_i).$ From \eqref{cjan13d} we obtain (after some manipulation) that if $k_{i+1}-k_{i-1}$ is small then
$$c_i\approx \frac{k_{i+1} - k_{i-1}}{2} \;u''(k_i), \quad i > 0. $$
In the case of $w_0 = u(0) = 0$ it also follows that  
\begin{equation*}
    c_0 = \frac{(w_1 - w_0)}{k_1} \approx u'(0).
    \label{cjan26a}
\end{equation*}

 \noindent For the target function $u(x) = x(1-x)$ for which the optimal set of approximating knots is uniform with spacing $h$, i.e.\ $h=k_{i+1}-k_i$, we have 
$c_0 \approx 1$ and $c_i \approx -  h$ for $i > 0$.
Observe that there is a large jump from $c_0$ to $c_1$ leading to a poor balancing of the \ReLU \,NN representation, while the FKS representation is better balanced.
 If, instead, we assume that the knot points have a density of $1/|u''|$ then the values of $c_i$ are approximately constant and  the \ReLU \,NN representation is evenly balanced. This observation links the coefficients of the \ReLU \,NN to the equidistribution condition and we investigate this further in Section \ref{sec:loss_ifks}.

\section{Training the ReLU NN and the FKS}\label{sec:loss_training}

Having seen the expressivity of the \ReLU\,NN and of the FKS and the formal equivalence of the (shallow) \ReLU \,NN and the FKS approximation in Section \ref{sec:Notation}, in this and the following two sections we consider how to train these to find the best approximation of the given function $u$. 
In all cases the training will be done through a (first order) gradient-based approach using the Adam optimiser with a suitable initial start and with an appropriate loss function. This procedure is chosen as it is a very common way of training a machine learning network. We find in general that a direct training of all of the parameters is problematic, and that a more effective approach is to divide the training procedure into first training the knots/breakpoints and then the weights of the network. Different loss functions are used in these two cases, and preconditioning is needed for the \ReLU \,NN network. To unify notation, we denote  the \ReLU \,NN or the FKS approximation by $y(x)\colon [0,1]\to \mathbb R$ and the associated parameter values by $\mathbf{\theta}$. Further, we denote the target function by $u\colon [0,1]\to \mathbb R$.
We initially consider  the  loss function $\mathcal L_2^2$ based directly on the {\em mean-squared approximation error} of the approximation. This is the default loss function which is typically used in training a NN approximation.

\vspace{0.1in}

\noindent As both $y(x)$ and $u(x)$ are given for all $x \in [0,1]$ we have that the $\mathcal L^2_2$-error of the approximation is given by
\begin{equation}
\mathcal  L_2^2(\mathbf{\theta}) = \int_0^1 (y(x,\mathbf{\theta}) - u(x))^2 \; dx.
\label{eqn:l2}
\end{equation}
 In practice, we cannot evaluate \eqref{eqn:l2} for a general function and we have to consider various approximations to it. For this, we consider  a set of {\em quadrature points} $\{x_k\}_{k=0}^{s-1}\subset [0,1]$ for some large parameter $s\in \mathbb N\backslash\{0\}$  which can either be randomly generated with uniform distribution or regularly distributed over the domain $[0,1]$. 
Assume that points $\{x_k\}_{k=0}^{s-1}\subset [0,1]$  are ordered so that $x_i < x_{i+1}$. An approximation to $\mathcal L_2^2$ in \eqref{eqn:l2} is  given by
 \begin{equation}
L_{2}^2(\mathbf{\theta}) = \sum_{i=0}^{s-1} (x_{i+1}-x_i) \Big( y(x_i,\mathbf{\theta}) - u(x_i)\Big)^2. 
\label{eq:L2_loss_function}
\end{equation}  
Note that minimising \eqref{eq:L2_loss_function} is a highly non-convex problem, and the minimisation process generally leads to sub-optimal solutions. 
A similar loss function is often used in the {\em Deep Ritz Method (DRM)} \cite{E_DeepRitz}. Similarly to the DRM,
 we will consider both the case where $\{x_k\}_{k=0}^{s-1}$ are sampled at each iteration or fixed during the training. 
Observe that the  minimisation of $L_{2}^2$ in \eqref{eq:L2_loss_function} reduces to the pointwise error  at  sample points $\{x_i\}_{i=0}^{s-1}$ and neglects global properties of the function, such as its curvature. This is important when considering equidistributed knots in regions of highest curvature, which  contribute significantly to the approximation error. 

\vspace{0.1in}

\noindent Whilst simple and widely used, for example in general \ReLU \,NN training or in linear spline approximation, the use of the loss function (\ref{eq:L2_loss_function}) {\em is problematic} when used to train either a \ReLU \,NN or an FKS network. We can see this clearly in the results presented in Figure 1 where we see poor accuracy of both the FKS and \ReLU \,NN networks trained using this loss function. Similar behaviour can be seen in many other examples and we will give a series of numerical calculations giving evidence for this is Section 8.

\vspace{0.1in}

\noindent There are a number of reasons for this poor behaviour. Firstly all of the terms in the parameter set $\mathbf{\theta}$ are treated equally, whereas the knots/breakpoints $k_i$ play a very different role in the approximation from the weights $w_i, c_i$. This leads to a highly non-convex minimisation problem. Secondly, in the the case of the \ReLU \,NN approximation it leads (as we shall show) to an ill-conditioned problem for the weights. 

\vspace{0.1in}

\noindent Accordingly we now consider an alternative procedure for finding the best approximation which takes into account the distinguished role played by the knots/breakpoints. This firstly solves the nonlinear problem of finding a close approximation to the knots and then the nearly linear problem of finding the wights and the optimal knots of the network. To do this we make use of the interpolating FKS

\section{Finding the optimal knots of the interpolating FKS using equidistributon based loss functions}\label{sec:loss_ifks}

\noindent Whilst the interpolating FKS (IFKS) $\Pi_1 u$ is sub-optimal as a general approximating function, it is an excellent first guess for  general optimisation procedures. Finding the IFKS only involves determining the knots $k_i$ for $i=1,\ldots,N-1$ which leads to a  simplified   approximation process approach for choosing the optimal knots.  We show presently that this process can  be adapted for training the breakpoints of the ReLU \,NN. Having found the knots, the problem of finding the weights of the optimal FKS is a well conditioned nearly linear problem and we consider this in the next section together with the problem of finding the coefficients of the ReLU \,NN.

\vspace{0.1in}

\noindent To find the optimal knots for the IFKS we consider optimising a different loss function from the standard $L_2$ loss above. This loss function is based on the concept of equidistributing the IFKS solution error across the knot locations.

\subsection{The equidistribution loss function}
 
\noindent Provided that the target function $u(x)$ is twice differentiable in $(k_i,k_{i+1})$,  the  local interpolation error  of the  linear interpolant $y$ of $u$ on $[k_i, k_{i+1}]$ satisfies
\begin{equation}
    y(x,\mathbf{\theta}) - u(x) \approx \frac{1}{2} (x-k_i)(k_{i+1}-x) u''(x_{i+1/2})
    \label{ca1}
\end{equation}
for any  $x\in [k_i, k_{i+1}]$ and some $x_{i+1/2} \in (k_i,k_{i+1})$.
Note that this result even holds for the functions such as  $u(x) = x^{2/3}$ for which $u''(0)$ is singular. 
This yields 
\begin{equation*}
    \int_{k_i}^{k_{i+1}} (y(x,\mathbf{\theta}) - u(x))^2 \; dx \approx 
    \frac{1}{120} (k_{i+1} - k_i)^5  |u''(x_{i+1/2})|^2,
\end{equation*}
implying
\begin{equation}
 \mathcal L_2^2(\mathbf{\theta})=   \int_0^1 (y(x,\mathbf{\theta}) - u(x))^2 \; dx \approx
    \frac{1}{120} \sum_{i=0}^{N-2} (k_{i+1} - k_i)^5   |u''(x_{i+1/2})|^2.
   \label{ca3.5}
\end{equation}

\noindent  Motivated by \eqref{ca3.5}, we define
\begin{equation} 
L_I(\mathbf{\theta}) = \frac{1}{120} \sum_{i=0}^{N-2} \; (k_{i+1} - k_i)^5  \; |u''(k_{i+1/2})|^2 
\label{ca4}
\end{equation}
where $k_{i+1/2}=k_i+k_{i+1}$ and we obtain $\mathcal L_2^2(\mathbf{\theta})\approx L_I(\mathbf{\theta})$.
Note that $L_I$ only depends  on the free knot locations for the interpolating FKS, and thus can be used to train the knots. However, unlike other loss functions $L_I$ requires knowledge of $u''$ rather than point values of $u$ and we will assume for the moment that $u''$ is known. 

\vspace{0.1in}

\noindent A  powerful result in the form of the equidistribution principle, first introduced by de Boor \cite{deBoor1973} for solving boundary value problems for ordinary differential equations,  gives a way of finding $\mathbf{\theta}$ so that $L_I$ in \eqref{ca4} is minimised. 
More precisely, minimising $L_I$ in \eqref{ca4} can be expressed by equidistributing the error over each cell which results in the following lemma (see \cite[Chapter 2]{HuangRussel}):

\vspace{0.2in}

\noindent {\bf Lemma 5.1 [Equidistribution]}\label{lem:equidistribution}
 {\em The loss $L_I$ in \eqref{ca4}
    is minimised over all knots $k_i$ if 
    \begin{equation}\label{eq:rho}
        \rho_{i+1/2}= (k_{i+1} - k_i)  \; |u''(k_{i+1/2})|^{2/5} \quad \text{for all }i=0,\ldots,N-2
    \end{equation}
    there is a constant $\sigma>0$ so that 
    \begin{equation}
   \rho_{i+1/2}= \sigma\quad \text{for all }i=0,\ldots,N-2.
    \label{ca6}
    \end{equation}
}

\vspace{0.1in}

\noindent The equidistribution principle in Lemma \ref{lem:equidistribution} replaces a non-convex optimisation problem for finding all of the terms of ${\mathbf{\theta}}$ by one of just finding the knots ${\mathbf k}$ with better convexity properties.   A set of knot points satisfying these conditions is said to be {\em equidistributed}. The degree of equidistribution can then be used as a quality measure of the solution.

\vspace{0.1in}

\noindent The algebraic equations \eqref{ca6} can in principle be solved directly \cite{deBoor1974}, or iteratively  using   moving mesh techniques such as \cite{Huang1994} for example. We propose instead training the knots of the IFKS directly using an optimisation approach enforcing the equidistribution condition \eqref{ca6} directly through minimising an equidistribution-based loss function. We do this using Adam with an initially uniform distribution of the knots as a start. We demonstrate that this procedure is both effective, and can be generalised to the training of the \ReLU \,NN.

\vspace{0.1in}

\noindent  To do this, we compute $\rho_{i+1/2}$ from  knots $k_i$ by \eqref{eq:rho} and set $\sigma$ as their mean, i.e.\ $\sigma = \frac{1}{N-1} \sum_{i=0}^{N-2} \rho_{i+1/2}.$ We define the equidistribution loss function $L_E$ by 
\begin{equation}
    L_E (\mathbf{\theta})= \sum_{i=0}^{N-2} (\rho_{i+1/2} - \sigma)^2.
    \label{coct11}
\end{equation}

\vspace{0.1in}

\noindent {\bf Remark 5.2} In practice, to ensure regularity  when $u''$
is small we replace the definition of $\rho_{i+1/2}$ in  (\ref{eq:rho}) by the regularised version
\begin{equation*}
        \rho_{i+1/2}= (k_{i+1} - k_i)  \; (\epsilon^2+u''(k_{i+1/2}))^{1/5} \quad \text{for all }i=0,\ldots,N-2.
    \end{equation*}
A value of $\epsilon^2 = 0.1$ works well in practice, and we will use it for further calculations.

\subsection{Convergence of the IFKS}

To show why it is important to first find the knots of the IFKS we show the excellent accuracy ${\mathcal O}(1/N^4)$ of this approximation. The optimal knots $k_i,~ i=0,\ldots,N-1$, of $L_I$ for $\Pi_1 u$ are given analytically by using quadrature applied to (\ref{eq:rho}) in the one-dimensional setting, and we will consider this approach in Section 8.3. 
Similar to Moving Mesh PDEs \cite{HuangRussel}, we consider the {\em physical interval} $x \in [0,1]$ to be a map from a {\em computational interval} $ \xi \in [0,1]$ so that $x = x(\xi)$ with $x(0) = 0$ and $x(1) = 1$.
For $i = 0, \ldots, N-1$, the $i th$ knot point $k_i$ is  given by
$k_i = x(i/(N-1))$.
Provided that $N$ is large we can then use the approximations
$$k_{i+1} - k_i = \frac{1}{N-1}\frac{d x}{d \xi}( \xi_{i+1/2})\quad \text{and}\quad x_{i+1/2}=x(\xi_{i+1/2})$$
for some $ \xi_{i+1/2} \in (i/(N-1),(i+1)/(N-1))$, where $x_{i+1/2} \in (k_i,k_{i+1})$ is defined by \eqref{ca1}.
The equidistribution condition  for $L_I$   requires that $(k_{i+1} - k_i)^5 |u''(x_{i+1/2})|^2$ is constant for $i=0,\ldots, N-1$ which yields
\begin{equation}
    \left(   \frac{dx}{d \xi}( \xi_{i+1/2}) \right)^5 \left(u''(x(\xi_{i+1/2}))\right)^2 = D^5
    \label{coct21}
\end{equation}
for a suitable constant $D$. This results in a differential equation for $x$ given by
\begin{equation}
    \frac{dx}{d \xi}(\xi)  = D \; \left(u''(x(\xi))\right)^{-2/5}, \quad x(0) = 0, \quad x(1) = 1,
   \label{nov15a}
\end{equation}
where value of $D$ is fixed by the boundary condition so that
\begin{equation}
    D = \int_0^1 (u''(x(\xi)))^{2/5} \; d\xi.
    \label{nov15b}
\end{equation} 
The optimal knots $k_i = x(i/(N-1))$ of a  piecewise linear IFKS on an equidistributed mesh can then be determined from the solution of  \eqref{nov15a} using \eqref{nov15b}.
We substitute \eqref{coct21} into \eqref{ca4} to obtain 
\begin{equation*}
    L_I(\mathbf{\theta}) \approx \sum_{i=0}^{N-2} \left( \frac{dx}{d \xi}( \xi_{i+1/2}) \right)^5\frac{(u''(x(\xi_{i+1/2})))^2}{120 (N-1)^5} = \sum_{i=0}^{N-2} \frac{D^5}{120 (N-1)^5} = \frac{D^5}{120 (N-1)^4}
\end{equation*}
up to leading order.
We deduce that provided $D = {\mathcal O}(1)$, the discretisation $L_I$ of $\mathcal L^2_2$ for an interpolating FKS is then ${\mathcal O}(1/N^4)$ as required.

\vspace{0.1in}

\noindent As an  example, we consider  the target function 
$u(x) = x^{\alpha}$ for some $ \alpha \in(0, 1)$ in the supplementary material \ref{ex:optimal}.

\subsection{Training the breakpoints of the ReLU \,NN}

This procedure can be applied directly to determine the breakpoints of the \ReLU \,NN. This follows immediately from the formal equivalence of the \ReLU \,NN and the FKS with knots $k_i$ given indirectly by (3.1) We simply train for the $k_i$ using the methodology described earlier in this section. 

\section{Training the coefficients of the shallow ReLU \,NN and  FKS representations}
\label{sec:conditioning}

\subsection{Overview}

\noindent Having found the optimal knots of the interpolating FKS we now optimise the ReLU \,NN/FKS representations given by (\ref{cnov123}). We find empirically that the best knots for the FKS are very close to those of the IFKS. For a given set of {\em fixed} knots this problem then reduces in the case of spline approximation to that of the well-conditioned least minimisation problem of finding the best coefficients $w_i$. In contrast the determination of the best coefficients $c_i$ for the \ReLU \,NN is (as we have seen in Figure 2, and will now prove) is an ill-posed least squares problem with very slow convergence of the optimisation procedure. However, by a simple preconditioning step we can transform the \ReLU \,NN network into one which can be readily optimised to accurately approximate complex functions. Accordingly, we now consider the problem of using the $L_2$ loss function to determine $w_i$ and $c_i$ for known $k_i$ (assumed to have been ore calculated using the IFKS).

\subsection{Conditioning of the training}

 Using the results of Section 3 we compare the conditioning of the problem of calculating the scaling factors $c_i, i=0 \ldots N-2$ and the weights $w_i, i=1 \ldots N-1$ of the shallow \ReLU \,NN and the FKS representations respectively (to give the best $L_2$ approximation of the function $u(x)$) on the assumption that the knot locations $k_i$ are known. We show that, as expected, the FKS problem (with the localised support basis functions) is well conditioned for all $N$ whereas the \ReLU \,NN problem (for which basis functions have global support) is increasingly ill-conditioned for larger values of $N$.
In particular we establish the following

 \vspace{0.1in}

 \noindent {\bf Lemma 6.1} {\em For fixed knots $k_i$, $i = 1 .. N$, the normal equations for finding the coefficients $w_i$ of the FKS have condition number $\kappa = {\mathcal O}(1)$. In contrast the normal equations for finding the coefficients $c_i$ of the \ReLU \,NN have condition number $\kappa = {\mathcal O}(N^4)$}

\vspace{0.1in}

\noindent To prove this result we observe it follows from \eqref{cjan13d} that \begin{align}\label{eq:cvec}
   \boldsymbol c = A \boldsymbol w
\end{align}
for a linear operator $A$ and we can compare the conditioning of the two optimisation problems by studying the properties of the matrix $A$. 
The matrix $A$ has a specific structure which we can exploit. It follows from (\ref{cjan13d}) that
\begin{align}\label{eq:coeffeq}
    c_i = \frac{w_{i+1} - w_i}{k_{i+1} - k_i} - \frac{w_i - w_{i-1}}{k_i - k_{i-1}}
\end{align}
for $i=2,\ldots,N-2.$
Hence,  $\boldsymbol c=(c_i)_{i=1}^{N-2}, {\mathbf w^*} = (w_i)_{i=1}^{N-2} \in \mathbb R^{N-2}$ are related via
\begin{equation}
{\mathbf c} = T {\mathbf w}^* + \gamma_{N-2} w_{N-1} {\mathbf e}_{N-2},
\label{cap1}
\end{equation}
where ${\mathbf e}_{N-2} = (0,0,0, \ldots 1)^T \in R^{N-2}$ and $T \in R^{(N-2)\times (N-2)}$ is the symmetric tri-diagonal matrix with ${\mathbf \beta}_i$ on the leading diagonal, and ${\mathbf \alpha_i}$ and ${\mathbf \gamma_i}$ on the upper and lower first diagonals. 
Observe that to find ${\mathbf w}$ from ${\mathbf c}$ (and hence to invert the linear operator $A$) we can let $w_{N-1}$ be initially unknown, invert $T$ using the Thomas algorithm, to find ${\mathbf w}^*$ in terms of $w_{N-1}$ and then fixing $w_{N-1}$ from the relation 
$w_1 = c_0/\alpha_1$.
Evidently the conditioning of the operator $A$ is completely determined by the conditioning of $T$.

\subsection{Uniformly spaced knots}

\noindent We can make very precise estimates of the conditioning of the problem of finding ${\mathbf w}$ and ${\mathbf c}$ in this case. Suppose that we have uniformly spaced knots $k_i = i/(N-1)$, then (\ref{eq:coeffeq})
reduces to 
$ c_i = (N-1)(w_{i+1} - 2w_i +  w_{i-1}).$
The linear operator $T$ then becomes the tri-diagonal Toeplitz matrix
\begin{align*}
T= (N-1)\begin{pmatrix}
-2 & 1 & 0 & \cdots & \cdots & \cdots & \cdots & 0\\
1 & -2 & 1 & 0 & & & & \vdots\\
0 & 1 & -2 & 1 & \ddots & & & \vdots\\
\vdots & 0 & \ddots & \ddots & \ddots & \ddots & & \vdots\\
\vdots & & & \ddots & 1 & -2 & 1 & 0\\
\vdots & & & & 0 & 1 & -2 & 1\\
0  & \cdots & \cdots   & \cdots & \cdots & 0 & 1 & -2\\
\end{pmatrix}\in \mathbb R^{N-2\times N-2}.
\end{align*}
We can then apply standard theory of $M \times M$ Toeplitz matrices of the tri-diagonal form taken by $T$. This theory implies that the eigenvalues of $T$ are given by 
$$\lambda_k =-2(M-1) + 2(M-1)\cos(\pi k/(M+1))\in (-4 (N-1),0)$$
for $k=1,\ldots,M$. In particular, we have $|\lambda _1| \leq \ldots \leq |\lambda _M|$, and for large $M$ we  have
\begin{align*}
\lambda_1 &= -2(M-1) +2(M-1) \cos\left(\frac{\pi}{ M+1}\right)\\&\approx -2 (M-1)+2(M-1)\left( 1-\frac{\pi^2}{2(M+1)^2}\right) = -\frac{\pi^2(M-1)}{(M+1)^2}.
\end{align*}
Similarly, for large $M$,
\begin{align*}
    \lambda_{M}  &= -2(M-1) +2(M-1) \cos\left(\frac{M\pi}{ M+1}\right)
    \approx -4(M-1),
\end{align*}
implying that the condition number $\kappa(T)$ satisfies 
\begin{align}\label{eq:condT}
\kappa(T)=\left|\frac{\lambda_M}{\lambda_1}\right|\approx \frac{4 (M+1)^2}{\pi^2}+1, \quad \mbox{with} \quad M = N-2.
\end{align}
For the FKS representation of the approximation to the target function, we aim to solve the problem for the weights ${\mathbf w}$ given by
 $ M \boldsymbol w =  \boldsymbol u$,
where $M_{i,j} = \langle \phi_i, \phi_j \rangle$, giving
 \begin{align*}
M= \frac{1}{6(N-1)} \begin{pmatrix}
4 & 1 & 0 & \cdots & \cdots & \cdots & \cdots & 0\\
1 & 4 & 1 & 0 & & & & \vdots\\
0 & 1 & 4 & 1 & \ddots & & & \vdots\\
\vdots & 0 & \ddots & \ddots & \ddots & \ddots & & \vdots\\
\vdots & & & \ddots & 1 & 4 & 1 & 0\\
\vdots & & & & 0 & 1 & 4 & 1\\
0 & \cdots & \cdots  & \cdots & \cdots & 0 & 1 & 4\\
\end{pmatrix}\in \mathbb R^{N\times N}.
\end{align*}
This is also a tri-diagonal Toeplitz matrix. Hence, by the standard theory it has eigenvalues given by
$$6(N-1) \mu_k =4 + 2\cos(\pi k/(N+1))\in (2,6) \quad k=1,\ldots,N.$$
Note that $\mu_N \leq \ldots \leq \mu_1 $ with  $6(N-1) \mu_N  > 2$, $6(N-1) \mu_1 < 6$ and condition number $\kappa(M) = \mu_1/\mu_N < 3$. As $N \to \infty$ we have $6(N-1) \mu_N\to 2$, $6(N-1) \mu_1 \to 6$ and 
$\kappa(M) \to 3.$  
The  normal equations for ${\mathbf w}$ are given by 
$M^T  M \boldsymbol w = M^T \boldsymbol u$.
The boundedness of $\kappa(M)$ in the limit of large $N$ lies at the heart of the regularity, and eases the construction  of the FKS approximation from the normal equations. We now consider the related problem of finding the coefficients ${\mathbf c}$ of the \ReLU\,approximation.  This is of course closely linked to the problem of finding the FKS coefficients ${\mathbf w}$. For simplicity we will consider the case where $w_0 = w_{N-1} = 0$ so that the operators $L$ and $T$ are the same. From \eqref{eq:cvec}, we then obtain 
$M^T  M T^{-1} \boldsymbol c = M^T \boldsymbol u$
and multiplication by $(T^{-1})^T$ yields
$\tilde M^T \tilde M  \boldsymbol c = \tilde M^T \boldsymbol u$
for the matrix 
$\tilde M = M T^{-1}$. To calculate the condition number of the matrix $\tilde M$ we note, that as the matrices $T$ and $M$ have exactly the same tri-diagonal Toeplitz structure, it follows from standard theory that they have {\em identical eigenvectors} ${\mathbf e}_k$, $k = 1 \ldots N$
with the eigenvector ${\mathbf e}_k$ having corresponding eigenvalues $\lambda_k$ and $\mu_k$ for the matrices $T$ and $M$ respectively. It follows immediately, that the eigenvectors of the matrix $\tilde{M}$ are also given by the vectors ${\mathbf e}_k$ with corresponding eigenvalues
$\nu_k = \mu_k/\lambda_k.$
Observe that for large $N$ we have $\nu_1 \to -6N/\pi^2$ and $\nu_N \to -1/2N.$ Hence we deduce that as $N \to \infty$, 
$\kappa(\tilde{M}) = \nu_1/\nu_N \to  12 \; N^2/\pi^2.$
 In particular, this shows that $\kappa(\tilde M) = \mathcal O(N^2)$ for $N\gg 1$. When solving the normal equations for the weights ${\boldsymbol c}$ of the \ReLU\,network, we must consider the condition number $\kappa$ of the matrix $\tilde{M} \tilde{M}^T$ which from the above calculation satisfies  $\kappa = {\mathcal O}(N^4)$. This proves Lemma 6.1 in this case.

\vspace{0.1in}
 
 \noindent The value of $\kappa$ for the \ReLU\,optimisation problem is very large even for moderate values of $N$ such as $N = 64$, and is huge compared with the ${\mathcal O}(1)$ condition number for the FKS approximation. Intuitively this is so large because the integral of the overlap of two \ReLU functions is a quadratic function and thus its integral grows rapidly with the support size. 
 Classical optimisation methods (such as BFGS) have convergence rate dependent on 
$(\sqrt{\kappa} - 1)/(\sqrt{\kappa} + 1) = 1 - {\mathcal O}(1/N^2).$
This leads to very slow convergence for large $N$. We see presently that the performance of the Adam optimiser is similarly very slow on this problem.

\subsection{Non-uniformly spaced knots}

\begin{figure}[htbp]
  \centering
   \includegraphics[width=0.6\textwidth]{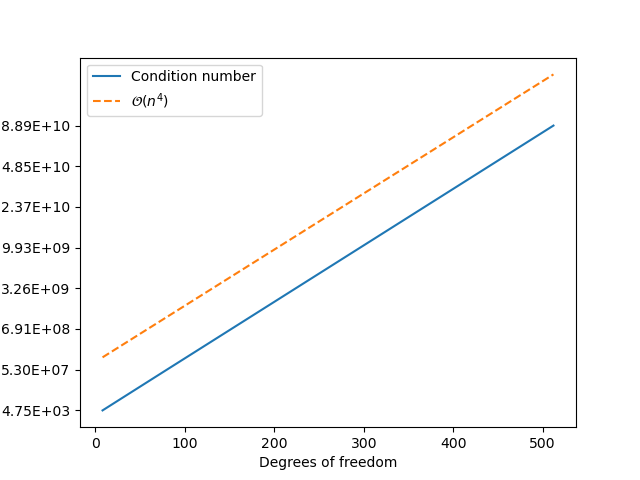}
   \caption{The condition number of the normal equations for the ReLU NN network of width $N$, showing that it is ${\cal O}(N^4).$}
    \label{fig:cdec1ff}
\end{figure}

 \noindent In this case the matrices $T$ and $M$ are still symmetric and tri-diagonal, but are not constant along their diagonals. 
 The (stiffness) matrix $M$ takes values $(k_{i+1} - k_{i-1})/3$ along its leading diagonal, and values $(k_{i}-k_{i-1})/6$ and 
$(k_{i+1}-k_{i})/6$ along its lower and upper diagonals, 
and thus is  diagonally dominant. A simple application of Gershgorin's circle theorem to estimate the location of its eigenvalues $\lambda_j$ implies that
$$\frac{1}{6} \min (k_{i+i} - k_{i-1}) \le  \lambda_j \le \frac{1}{2} \max (k_{i+i} - k_{i-1}).$$
Hence (consistent with the estimate on a uniform mesh)
$$\kappa(M) < 3 \; \frac{ \max (k_{i+i} - k_{i-1})} {\min (k_{i+i} - k_{i-1}).} $$
Note that this estimate for the condition number is smallest for  uniform knots, and increases if the knot spacing is non-uniform. 
In contrast, each row of the matrix $T$ has the respective coefficients:
$$\alpha_i = \frac{1}{k_i - k_{i-1}}, \quad -\beta_i = -\frac{1}{k_i - k_{i-1}} - \frac{1}{k_{i+1} - k_{i}}, \quad \gamma_i   = \frac{1}{k_{i+1} - k_{i}}.$$
Observe that $\alpha_i - \beta_i + \gamma_i = 0$ for each $i$. It follows that the matrix $T$ is near singular; indeed if ${\mathbf f} = (1,1,1,1,  ,\ldots 1)^T$ then 
$T {\mathbf f} = {\mathbf g} = (\gamma_1 - \beta_1, 0, \ldots, 0, \alpha_{N-2}- \beta_{N-2})^T.$
Hence 
$\|T^{-1}\| > \|f\|/\|g\| \to \infty$ as $N \to \infty.$
The condition number of $T$, and hence of $MT^{-1}$ thus increases without bound as $N$ increases. Numerical experiments  strongly indicate that, as in the case of the matrix $M$, the condition number increases more rapidly for non-uniform knots than uniform knots. In Figure \ref{fig:cdec1ff} we show the condition number for the normal equations used to optimise the coefficients of the ReLU NN for the test function $u(x) = \tanh(100(x-1/4))$ considered in Section 1, with the (non-uniform) breakpoints pre-trained to be the optimal values for the IFKS. We can see in this figure that the condition number scales as ${\mathcal O}(N^4) $ as expected

\section{Two-level training of  ReLU \,NN and FKS networks}\label{sec:equidistribution_training}

\noindent Given the results on the loss for the IFKS in Section \ref{sec:loss_ifks}, and the condition number estimates above, we consider two  procedures to train either a shallow \ReLU \,NN or a general FKS which combine the usual $L_2$ optimisation  with equidistribution and the calculation of the best IFKS. These lead to procedures which allow the expressivity of these methods to be better realised with close to optimal approximations.  We note that both of these methods require extra knowledge about the higher derivatives of the target function. We introduce the  {\em combined loss function} which combines the $L^2_2$-approximation error with the equidistribution condition and is given by
\begin{equation}
L_{\rm comb} \equiv L^2_2 + \beta L_E.
\label{eq:nov25}
\end{equation}
Observe that if $\beta$ is small the equidistribution condition acts to regularise the approximation error loss. Conversely when $\beta$ is large then the approximation error acts to regularise the equidistribution condition. The latter case is surprisingly effective. Often when invoking the equidistribution condition directly in adaptive methods it is found that the resulting mesh is irregular. Adding the approximation error indirectly as a regulariser appears to smooth the resulting knot location.

 \subsection{Method 1: Two-level training} In this approach we first find the knots/ breakpoints of the FKS/\ReLU \,NN. We then find the weights/scaling factors. This method is motivated by the observation whilst the interpolating FKS  $y(x) = \Pi_1 u(x)$ has quite a large error compared with the optimised solution, it is still quite close to the final solution, has a knot distribution very close to optimal, and has the correct asymptotic convergence properties. 

\vspace{0.1in}

\begin{enumerate}
    \item Suppose that we have either a FKS with knots $k_i$ given directly, or a shallow \ReLU \,NN with knots given indirectly by $-b_i/a_i$. We determine the knots $k_i$ of the Interpolating FKS $u(x) = \Pi_1 u(x)$ by  minimising the equidistribution loss (\ref{coct11}) for the  knots $k_i$. This is done by directly minimising the loss function $L_{\rm comb}$ using Adam with large $\beta$ (say $\beta = 10$). Typically  start with a uniform distribution of knots in the interval $[0,1]$.
    \item Use $\Pi_1 u(x)$ with the calculated knots as the {\em initial guess} for an optimising procedure based on the loss function $L^2_2$ in \eqref{eq:L2_loss_function} to find the coefficients of either the FKS or the shallow \ReLU \,NN. Or more simply, observing that the optimal knots for the FKS are very close indeed to those for the IFKS, freeze the knot locations and then solve the simple linear problem of finding the optimal weights $w_i$/scalings $c_i$.
\end{enumerate}

\vspace{0.1in}

 \noindent We see presently that this method applied as above is effective and well-conditioned in giving an expressive FKS approximation when using  an optimisation method such as Adam.
 In contrast when applied to the problem of finding the \ReLU \,NN coefficients, we find that whilst part (i) of the two-level training correctly locates the knot points $k_i$, applying part (ii) is still very slow for large $N$ due to  ill conditioning of the problem of finding the scalings $c_i$ identified in Section \ref{sec:conditioning}. This is an inherent feature of the \ReLU \,NN architecture we are considering. A simple resolution for this stage of the optimisation is to precondition the problem of finding the \ReLU coefficients $c_i$ by applying the transformation (\ref{eq:cvec}) after step (i). Hence, given a \ReLU \,NN we consider
{\em preconditioned two-level training of a \ReLU \,NN:}

\vspace{0.1in}

\begin{enumerate}
    \item Fix ${\mathbf c}$ and use equidistribution based loss to find $k_i$
    \item Apply $T^{-1}$ to determine ${\bf w}$ from ${\bf c}$ using the procedure outlined in Section 6 in which  we  find $T^{-1}$ efficiently in ${\mathcal O}(N)$ operations using the well known {\em Thomas Algorithm}. This preconditions the shallow \ReLU \,NN optimisation problem to that of finding the optimal FKS coefficients $w_i,k_i$.
    \item Apply (ii) to locate these coefficients.
    \item Finally we transform back to find $c_i$ from $w_i$ using (\ref{eq:cvec}). As the matrix $L$ is tri-diagonal the operation of multiplication by $L$ is ${\mathcal O}(N)$. 
\end{enumerate}

  \subsection{Method 2: Combined training}
 As an alternative, we consider a combined approach by using the loss function $L_{\rm comb}$ with a smaller $\beta$, say $\beta = 0.1$. This procedure can again be used for a shallow \ReLU \,NN or an FKS with some suitable initial conditions. For training the FKS, we use the loss function $L_{\rm comb}$ with the direct definition of $k_i$, while for training the NN we use the loss function $L_{\rm comb}$ with $k_i$ implicitly defined by \eqref{c1}. In the computations presented in Section 8, we find that this method works quite well for training the FKS (though not as well as the two-level method), but requires a careful choice of the regularisation parameter $\beta$. In practice a value of $\beta \approx 0.1$ seems to be appropriate for calculations. For the \ReLU \,NN problem it suffers from the same ill conditioning problems as the two-level training method, so preconditioning (as above) must be applied for it to be effective.


\section{Numerical Results}
\label{sec:Numerical Results}

\subsection{Overview} 
 We now give a series of numerical calculations for the training  and convergence   of  shallow univariate \ReLU \,NNs and different linear spline approximations, including the FKS. These calculations support the theoretical results of the previous sections, and confirm the viability of the algorithms we propose. The calculations are structured to first demonstrate the problematic issues with use of the 
 standard $L_2$ loss based training for the \ReLU \,NN architecture. We then look at the optimal case where we solve the equidistribution equation for the knots accurately using quadrature, and then find the weights of the optimal FKS. This gives a 'gold standard' for the accuracy which can be achieved using an adaptive linear spline approximation. We then compare the results from both of these sections with those achieved for the \ReLU \,NN and FKS using a variety of optimisation methods, including the two-level method described in Section 7. 
 
 \vspace{0.1in}

 \noindent We do this by considering the approximation of a set of target functions using the different loss functions and training procedures described earlier. To make the comparisons between the FKS approximations and the \ReLU \,NN comparable we assume that we have $N$ knots $k_i$ for $i=0,\ldots,N-1$, with $k_0 = 0$ and $k_{N-1} = 1$.  Accordingly, we consider the approximation of five different  target functions $u_i(x)$ for $u(x)$ on the interval $[0,1]$ of increasing complexity: 
\begin{align*}
    u_1(x) &= x(1-x), \quad
    u_2(x) = \sin(\pi x), \quad
    u_3(x) = x^{2/3}, \\
    u_4(x) &= \tanh(100(x-0.25)),\quad  u_5(x)= \exp(-500 (x-0.75)^2 ) \sin(20\pi x).
\end{align*}
Note that the first two target functions are very smooth, the third is singular as $x \to 0$, and the last two examples represent a smoothed step function and an oscillatory function, respectively, both with a mix of small and large length scales.
Our interest will be in the importance and speed of the training, the role of the initial start, and the convergence of the resulting approximation as $N \to \infty.$
Our extensive experiments on different optimisation methods such as Adam, Gauss-Newton   and Nelder-Mead optimisers all led to similar optimal solutions, computation times and training behaviour. Consequently, we focus on the Adam optimiser  for the remainder of this section. 

\vspace{0.1in}

\noindent As a summary of the results of this section, we find for all the examples, that with the usual $L_2$ loss (\ref{eq:L2_loss_function}) that the \ReLU\,trains slowly, if at all, leading to poor accuracy in the approximation. This is due to the strong non-convexity of the loss function, and the 
ill-conditioning problems identified in Section 6. Better, but still sub-optimal results are obtained when training the FKS with this loss function. In contrast, optimal results, with high accuracy, are obtained when training a FKS with the equidistribution-based loss function $L_{\rm comb}$ and two-level training. Similar results are found when using the same method, combined with preconditioning, to train the \ReLU \,NN.

\subsection{Poor convergence and ill-conditioning of the \ReLU \,NN training}

\begin{figure}[htbp]
    \centering
    \includegraphics[width=\textwidth]{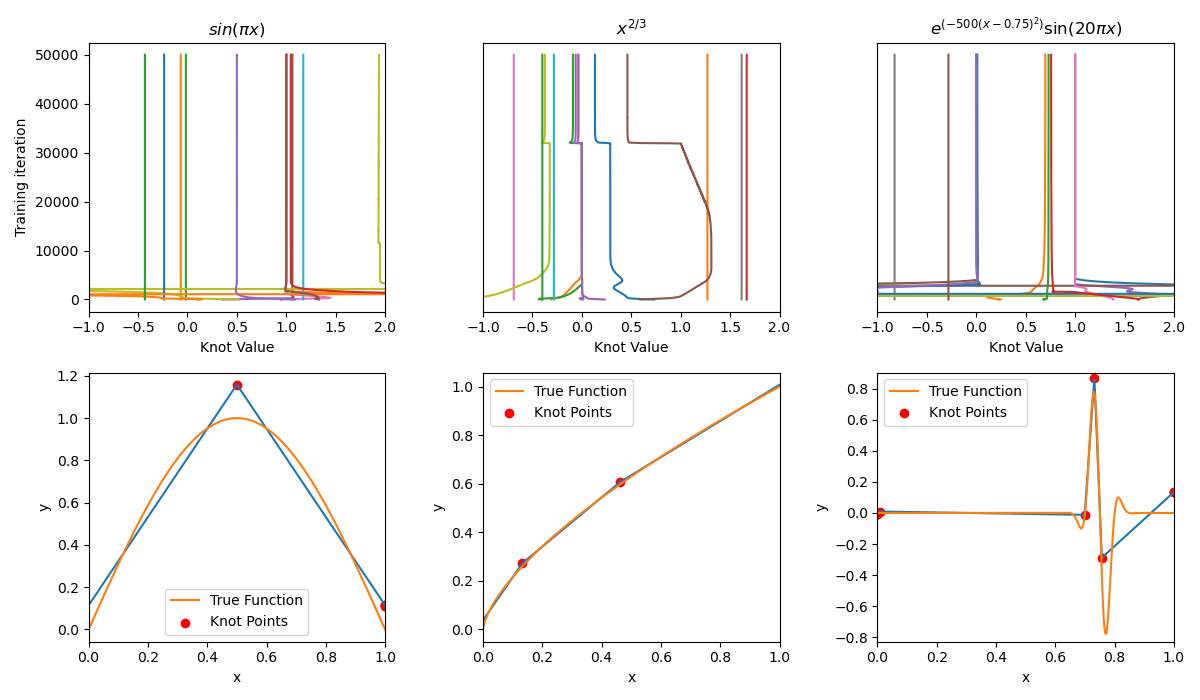}
    \caption{Comparison of (top) Knot evolution with standard Gaussian initialisation and (bottom) the trained approximation with the knot points indicated for different target functions.}
    \label{fig:sfu1}
\end{figure}

\noindent In Section 4 we claimed that the training of \ReLU \,NN with the standard $L_2$ loss function led to poor results, and in Section 6 proved the ill-conditioning of  the training of the scaling coefficients $c_i$ of this representation. We now present numerical evidence for these statements. We consider first the training of a shallow \ReLU \,NN with the usual  $L_2^2$ loss function given in \eqref{eq:L2_loss_function}, with width $W = 16$, and consider the approximation of the target functions $u_i(x), i=2,3,5$.  For this calculation the default PyTorch parameter initialisation is used, and the location of the implicit knot locations over 50,000 optimisation iterations using Adam with a learning rate of $10^{-3}$ are illustrated in the top part of Figure~\ref{fig:sfu1}. The lower part of Figure~\ref{fig:sfu1} shows the quality of the function approximation, with the breakpoints/implicit knot locations plotted. We see   that during the course of the training procedure the knots can cross over,  merge and  leave the domain $[0,1]$. 

\vspace{0.1in}

\begin{figure}[htbp]
   \centering
   \includegraphics[width=\textwidth]{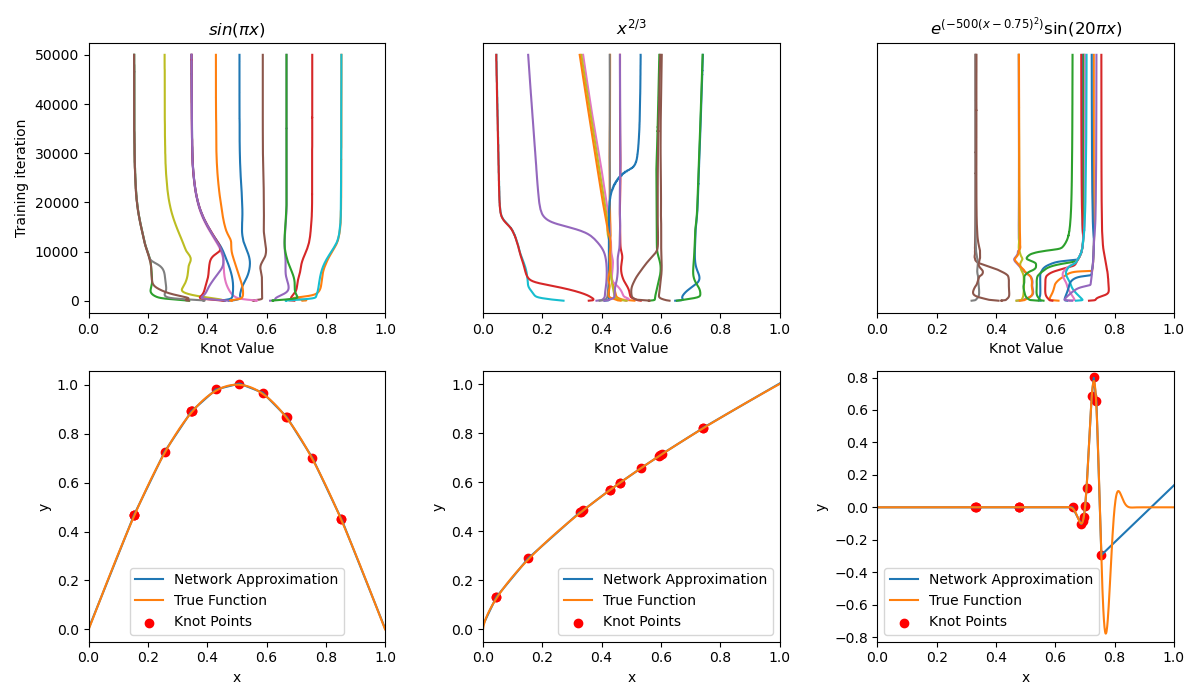}
   \caption{Comparison of (top) Knot evolution, with initialisation in $[0,1]$ and (bottom) the trained approximation with the knot points indicated for different target functions.}
    \label{fig:sfu2}
\end{figure}

\noindent  We next repeat the above calculation, but with the constraint on the starting parameters that the initial implicit knot locations given by \eqref{c1} are all required to lie in the interval $[0,1]$. The corresponding results are shown in Figure~\ref{fig:sfu2}. We can see that the results are better than those presented in Figure~\ref{fig:sfu1} but are still far from optimal, and show evidence of knot crossing and merging during training. 

\vspace{0.1in}

\noindent The resulting values of the $L^2_2$ loss function in the respective cases are:
$$\mbox{Random}: \quad u_2: 3.948\times 10^{-3}, \quad u_3: 2.611 \times 10^{-5}, \quad u_5: 8.524 \times 10^{-3},$$
$$\mbox{Constrained}: \quad u_2: 4.77 \times 10^{-6}, \quad u_3: 2.30 \times 10^{-6}, \quad u_5: 8.24 \times 10^{-3}.$$
We note that the loss values for the constrained start are in general   better than the   large loss values we obtain for the random start. 
However, we show presently that they are not close to those given by the FKS.

\vspace{0.1in}

\noindent  These calculations demonstrate  that the standard machine learning based techniques, when used in a simple example of functional approximation, can lead to a significantly sub-optimal solution in each of the three cases of the target functions. This can be seen by the distribution of the knots which is far from optimal. Note that they are irregular and unevenly spaced whereas the optimal knots would be expected to be symmetric, and regularly spaced. Furthermore the knots for the target functions $u_3$ and $u_5$ show no sign of clustering close to the singularity at $x=0$ (in the case of $u_3$) or the points of rapid variation (in the case of $u_5$). In all cases the training of the knots is erratic.  We conclude, as said in Section 4, that the commonly used training methods and loss functions for training a shallow \ReLU \,NN perform badly on even simple target functions. We now see how we can improve on this performance.

\vspace{0.1in}

 
  \noindent As proven in Section 6, the problem with training a \ReLU \, NN lies deeper than finding the optimal knots, and is also impacted by the poor conditioning of the problem of finding the associated coefficients $c_i$ even when the optimal knots for the IFKS are known. The case of taking $N=64$ and the target function $u_4(x)$ has already been illustrated in Figures 1 (right) and 2 where we use Adam with an equidistribution based loss function in the first stage of the two-level method to train both an FKS and a \ReLU \,NN to find the optimal knots $k_i$. Starting from these knot locations and with an initially random distribution of the scaling coefficients/weights, we then apply the second stage to further train both networks to find the optimal values of the coefficients $c_i$ for the \ReLU \,NN and $w_i$  for the FKS, respectively. It is clear from Figure~1 that the FKS trains much more quickly to a much more accurate approximation than the \ReLU \,NN for this target function. This is due to the poor conditioning of the problem for large $N$ associated with the condition number ${\mathcal O}(N^4)$ of the normal equations identified in Section 6 and in Figure 2. For example if $N =64$ then a direct calculation for this specific approximation problem gives 
$\kappa_{\ReLU} = 6.3 \times 10^7$ and $\kappa_{\rm FKS} = 3.7$. Similar poor conditioning is observed for all of the target functions considered.


\subsection{Optimal FKS approximations using MMPDEs}

 \noindent By way of contrast, we now show that very high approximation accuracy can be achieved for a correctly trained FKS obtained by using moving mesh PDE (MMPDE) based methods described in Section 5.2 to find the optimal knots, and then least-squares optimisation to find the weights. This gives a 'gold standard' result that can be used for comparison with optimisation based methods. To do this calculation we consider solving a regularised form of equations \ref{nov15a},\ref{nov15b}, and define a monitor function 
$m_j= (\epsilon_j + u_j''^2)^{1/5}$.
The value of the non-negative regulariser $\epsilon_j$ is not critical, but has to be chosen with some care when the value of $u_j''$ varies a lot over the domain to ensure that regions where $u_j''$ is relatively small still have some knots within them. We take $\epsilon_j = 0$ for $j=1,2$ and $\epsilon_j = 1$  for $j = 3,4$. From \ref{nov15a},\ref{nov15b} it follows that the optimal knot points $k_i, i=0, \ldots, N-1$, are given by $x(\xi)$ where $\xi = i/(N-1)$ and $x(\xi)$ satisfies 
\begin{equation}
\frac{dx}{d\xi} = \frac{\int_0^1 m_j(\tilde x) \; d \tilde x}{m_j(x)}, \; \quad x(0) = 0.
\label{cjan9a}
\end{equation}
To find the best  knot points $k_i$ for the  IFKS  we solve (\ref{cjan9a}) and evaluate the integral in the expression  by using a high-order Gear solver with a high tolerance. We then find the optimal weights by solving the $L_2$ minimisation problem. Note that for the target functions $u_1(x) = x(1-x)$ and $u_3(x) = x^{2/3}$ that the optimal IFKS knot points are given analytically (see the results in the Supplementary Material \ref{ex:optimal}) by $k_i = i/(N-1)$ and $k_i = (i/(N-1))^{15/7},$
respectively. 

\vspace{0.1in}

\begin{figure}[htbp]
\centering
        \includegraphics[width=0.95\textwidth]{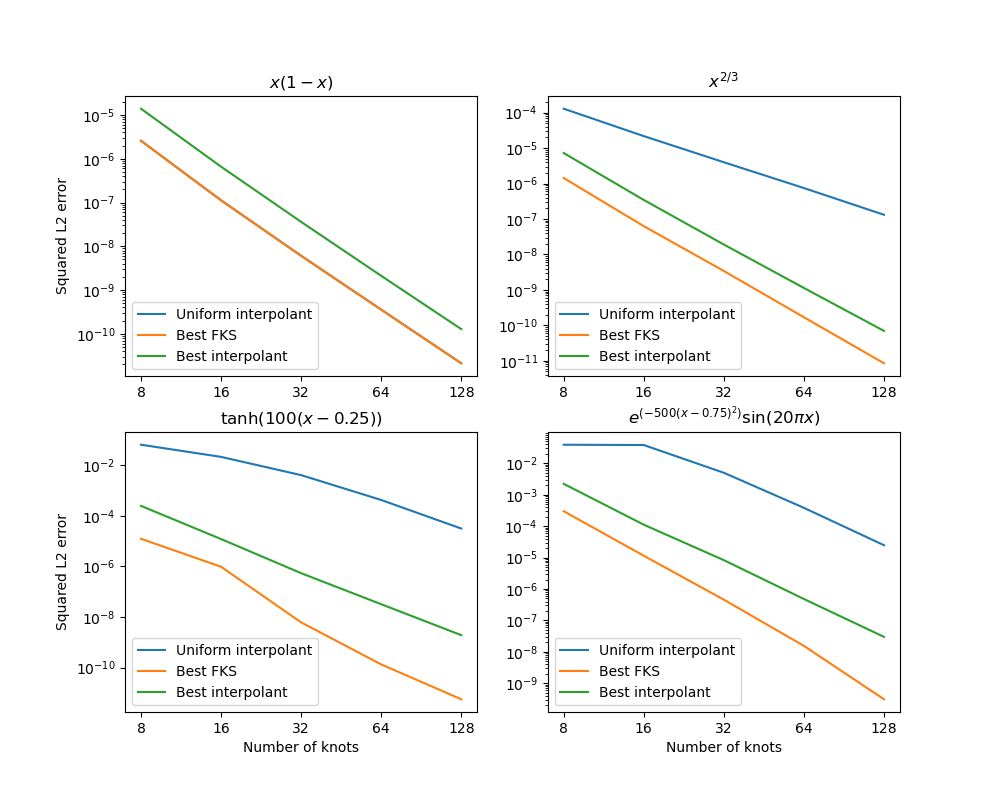}
    \caption{Comparison of the $L_2^2$ loss for linear spline approximations of four different target functions with different numbers of knots: blue - PWL interpolant on a uniform mesh, orange- optimal IFKS, green - optimal FKS}
    \label{fig:cjan10}
\end{figure}

\noindent  We present results for each of the target functions in Figure  \ref{fig:cjan10}.  
For each target function we study the convergence of the approximation by plotting the $L^2_2$ error as a function of $N$ in the three cases of (i) (blue) the PWL interpolant defined over a uniform mesh,  (ii) (orange) the optimal IFKS interpolant, and (iii) (green) the fully trained FKS. In each case the training was rapid. 

\noindent We see that the optimal IFKS and FKS both show ${\mathcal O}(1/N^4)$ convergence of the loss function.  The FKS gives the best results as expected, often significantly better than the optimal IFKS. In this context the 'optimal' FKS achieves the best expressivity of piecewise linear approximations which includes the \ReLU\,approximation, with errors of around $10^{-10}$ when $N = 64$. The error of the best $L_2^2$ PWL approximation on a uniform mesh is always much poorer, by several order of magnitude,  than the IFKS or FKS. But, as we have seen, it is much better than the naively trained \ReLU\,network. For the {\em smooth} target functions $u_1,u_2,u_4,u_5$ we see    ${\mathcal O}(1/N^4)$ convergence of the uniform PWL approximation for values of $N$ so that $h=1/N$ is smaller than the smallest length scale of the target function, but with a much larger constant of proportionality than either the optimal FKS or the optimal IFKS. In the case of the {\em singular} target function $u_3(x)$ we see  a slower convergence of the uniform PWL approximation at the theoretical rate of ${\mathcal O}(1/N^{7/3}).$

\subsection{\ReLU \,NN and FKS training using optimisation}\label{sec:num_results_equidistribution}

In this subsection, we present a set of numerical results 
for directly training a \ReLU \,NN, a linear spline (without free knots), and a  FKS for each of the target functions using an {\em optimisation approach}. In all case we use the Adam optimiser with a learning rate of $10^{-3}$. We consider the use of different loss functions, including  the equidistribution-linked loss function $L_{\rm comb}$  in  \eqref{eq:nov25} and using the  two-level training procedure  described in Section \ref{sec:equidistribution_training} where we first train the knots of the approximation and then the weights, in the case of the \ReLU \,NN with and without preconditioning.

\vspace{0.1in}

\noindent  We present the results as follows. For each target function $u_i(x)$ we consider using the following methods:
(a) {\em relu-regular}: A standard  \ReLU \,NN. This is the usual implementation of the \ReLU \,NN approximation in PyTorch as considered in the previous subsection, with random  initialisation of the Adam optimiser but with the knot locations constrained to lie in $[0,1]$.  (b) {\em relu-l2}: a \ReLU \,NN and {\em fks-l2}: a FKS trained using the standard $L^2_2$ loss function.  (c) {\em relu-2level}: a \ReLU \,NN trained in the  two-level manner described in Section 7.1 to first to locate the knots $k_i$ and then the weights {\em without  preconditioning}, {\em fks-2level}: a FKS trained using the two-level method which is equivalent to the ReLU NN trained using the two-level method {\em with preconditioning}, (d) {\em relu-joint}: a \ReLU \,NN and {\em fks-joint}: a FKS trained using the combined/joint  $L_2^2$ and equidistribution loss function described in Section 7.2 (without preconditioning). (e) {\em spline-regular}: the best least squares approximation, which takes a uniform knot distribution and optimises the weights.  Apart from case (a) we take the initialisation for the Adam optimiser to be the piecewise linear interpolant of the target function with uniform knots $k_i = i/(N-1), i = 0 \ldots N-1$ represented either in the form of the \ReLU \,NN or as a piecewise linear spline approximation. 

\vspace{0.1in}

\begin{figure}[htbp]
   \centering
   (a) Target function $u_1(x) = x(1-x)${
   \includegraphics[width=.98\textwidth]{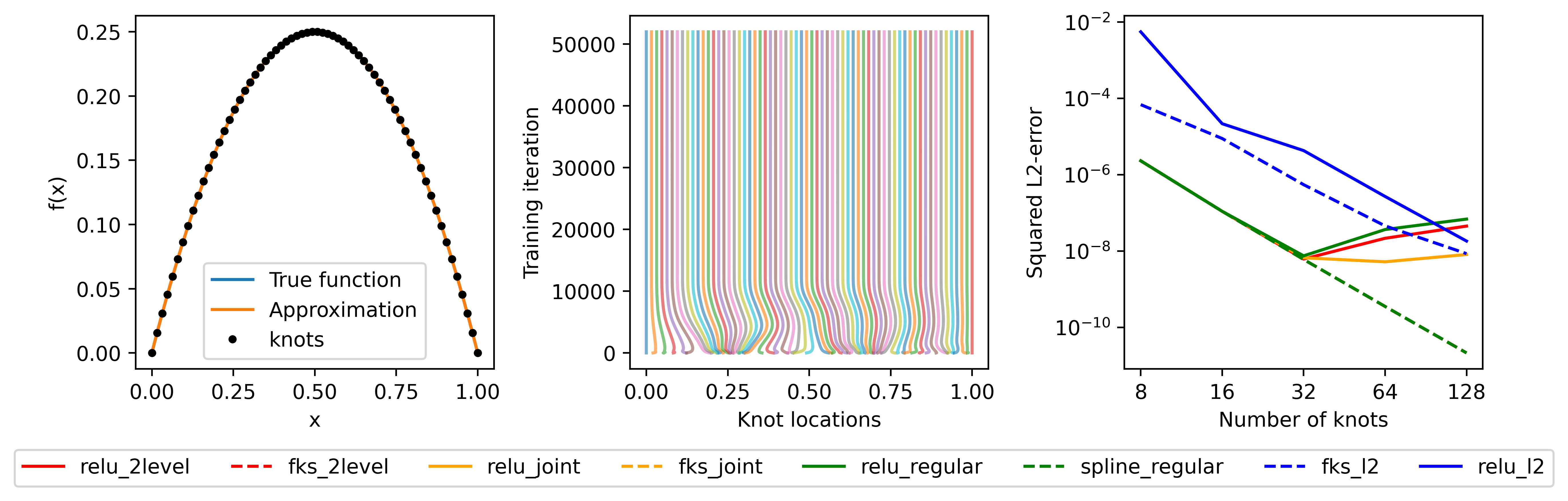}
   }\\
   (b) Target function $u_3(x) = x^{2/3}${
   \includegraphics[width=.98\textwidth]{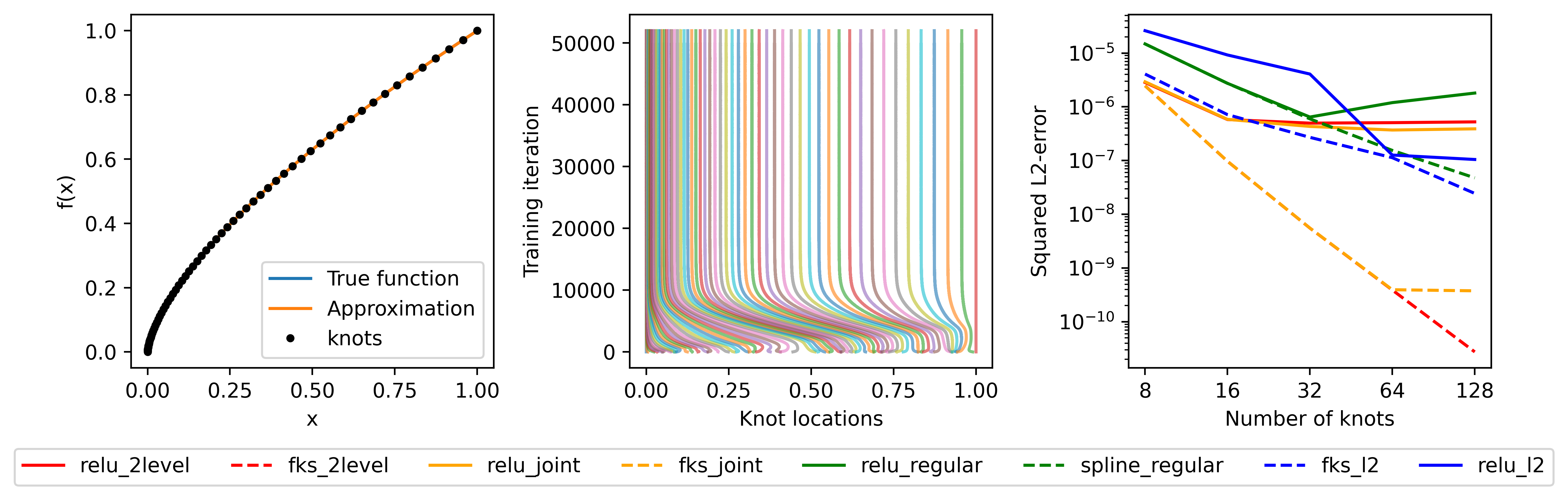}
   }\\
   (c) Target function $u_4(x) = \tanh(100(x-1/4))${
   \includegraphics[width=.98\textwidth]{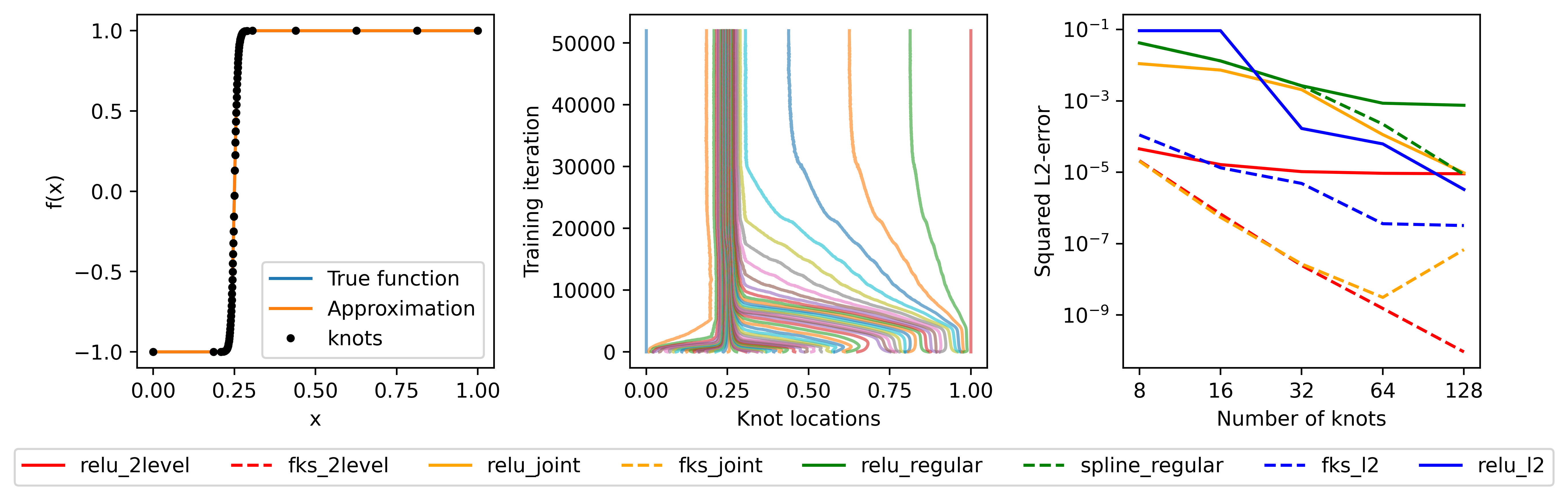}
   }\\
   (d) Target function $u_5(x) = \exp(500(x-3/4))^2) \sin( x).${
      \includegraphics[width=.98\textwidth]{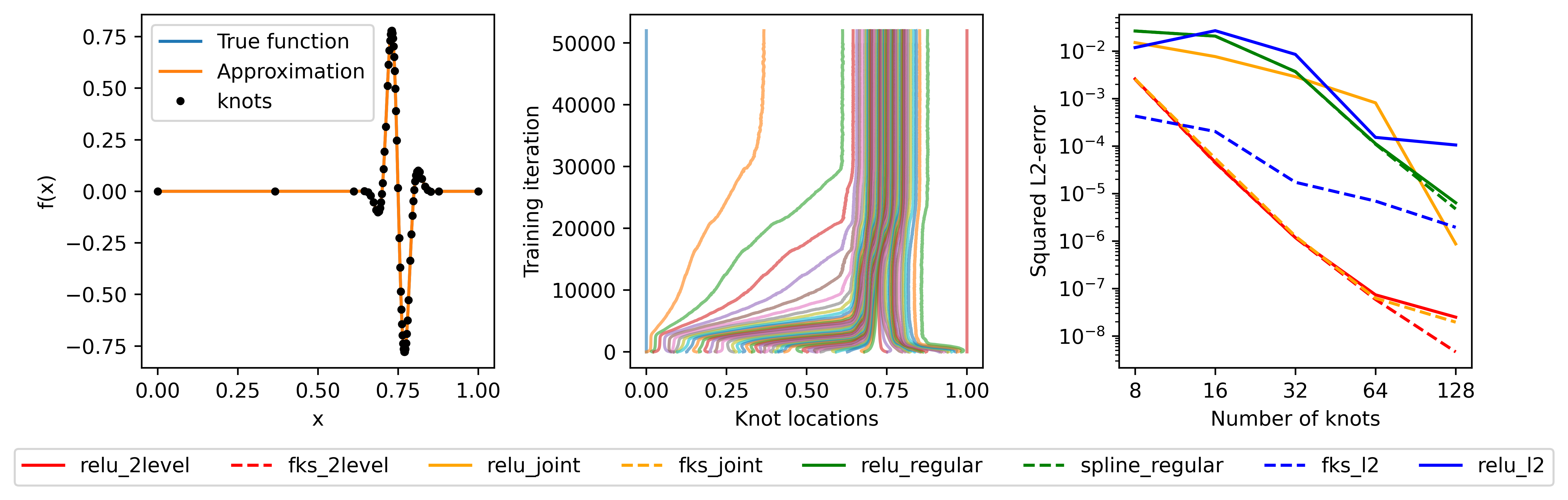}
      }
   \caption{Comparison of (i) Function approximation, (ii) knot evolution, (iii) Convergence for different target functions}
   \label{fig:cfeb1}
\end{figure}

\noindent  We show our numerical results for different target functions in Figure \ref{fig:cfeb1} in terms of different aspects: (i) The function approximation of the FKS trained using the two-level approach method when $N= 64$. (ii) The convergence of the knot values during training when using the equidistribution based loss function, where we plot (on the $x$-axis) the location of the knots as a function of time $t$ on the $y$-axis. This figure is nearly identical for both the FKS and the \ReLU \,NN approximations and the smooth evolution of the knots should be compared to erratic behaviour seen in Figures \ref{fig:sfu1} and \ref{fig:sfu2}. (iii) The convergence of the $L^2_2$ approximation error of each method as a function of $N$,

\vspace{0.1in}

\noindent The {\em basic \ReLU \,NN} trained using the direct $L^2_2$ loss function consistently performs badly. Whilst the approximation error does in general decrease with $N$, it is usually the worst of all the approximations by a significant margin. It also takes a much longer time to train than any other method due to the ill-conditioning of the system.

\vspace{0.1in}

 \noindent The {\em \ReLU \,NN trained using the direct two-level approach without preconditioning} in which the knots $k_i$ are first trained and then the weights are calculated {\em without} using preconditioning, works in general rather better than the basic \ReLU \,NN above, particularly for the lower values of $N$ where conditioning issues are less severe,  but is  outperformed by the FKS, particularly for larger $N$ where ill-conditioning issues arise. In the two-level approach the knot locations are found accurately in stage (i) of the training using the equidistribution based loss function and are presented in the middle figure.  Again, contrast this smooth evolution with the erratic convergence of the knots of the basic \ReLU \,NN presented earlier in Figures \ref{fig:sfu1} and \ref{fig:sfu2}.
However, having found the knot locations correctly, finding the resulting weights $c_i$ is ill-conditioned, as expected from the results of Section~5. Consequently the convergence to the correct solution is very slow, as was illustrated in Figure 1 (right).  

\vspace{0.1in}

\noindent  The {\em \ReLU \,NN trained using the direct two-level approach with preconditioning} behaves essentially identically to the FKS with two-level training, see later.

\vspace{0.1in}

 \noindent The {\em \ReLU \,NN} trained using the joint loss function behaves in a similar manner to the two-stage trained \ReLU \,NN without preconditioning.

\vspace{0.1in}

 \noindent The {\em regular classical least squares linear spline approximation on a uniform mesh of mesh spacing $h = 1/N$} so that there is no training of the knots, works well for the first two smooth target functions with convergence ${\mathcal O}(N^{-4}).$ For the singular target function $u_3(x)$ it converges but at the sub-optimal rate of ${\mathcal O}(N^{-7/3})$ described in the Supplementary Material \ref{ex:optimal}. (Exactly the same behaviour is seen for the piecewise linear interpolant on a uniform mesh.) In the case of the smooth target functions $u_4(x)$ and $u_5(x)$ we see sub optimal convergence when  $h = 1/N$ is greater than the smallest length scale of the problem (which is $1/100$ for $u_3$ and $1/\sqrt{500} = 1/22.3$. For larger values of $N$ (and smaller values of $h$) we see ${\mathcal O}(1/N^4)$ convergence. In all cases the error of the regular spline is much larger than that of the best FKS approximation, but in general better than that of the \ReLU \,NN without preconditioning.  

\vspace{0.1in} 

 \noindent The {\em full FKS (with free knots and weights) trained with either the joint loss function (knots and weights together with an equidistribution constraint) or the two-level training (first training the knots using equidistribution)} then the weights, is consistently the best performing method, alongside the near identical example of the {\em preconditioned two-level training of the \ReLU \,NN} with similar results in both cases. The results show excellent and consistent convergence at a rate of ${\mathcal O}(1/N^4)$. The convergence of the knots in all cases is regular and monotone, indicating a degree of convexity to the problem. The resulting knots are regularly spaced, showing symmetry and clustering where needed. The training converges rapidly in all cases. Note that pre-training the knots using the 'gold standard' MMPDE as discussed in Section 8.3 gives a slightly smaller error after optimisation than the best trained FKS/\ReLU \,NN (although the difference is not large). This is to be expected as the MMPDE has employed an accurate solver to determine the knot points.

\vspace{0.1in}

 \noindent The {\em FKS trained using the direct $L_2^2$ loss} function performs better than the remaining approximations, but a lot worse than the optimal FKS. We do not in general see ${\mathcal O}(1/N^4)$ convergence in this case.

\subsection{Conclusions from the results}


\noindent  A naively (in the sense of using the usual training procedures) trained shallow \ReLU \,NN, always gives a poor approximation of the target functions.  The knot points are generally badly placed, especially if we start from a random parameter set. Starting from a uniform set of knots gives an improvement, but still leads to a poor approximation. The training method is also very ill-conditioned. 

 \vspace{0.1in}

 \noindent Two-level training (using equidistribution to find the knots/breakpoints but without using preconditioning) of a \ReLU \,NN is significantly better than non-equidistribution based training, and the knot points are close to optimal locations. However the training of the weights $c_i$is far from optimal due to ill-conditioning for larger values of $N$, leading to an approximation that is still far from being as expressive as it could be. 

\vspace{0.1in}

 \noindent Two-level training of an FKS (using the optimisation approach) is better than non-equidistribution based training as it is faster and more accurate and gives a good approximation which is very expressive. Preconditioning the \ReLU \,NN in the two-level training (effectively turning it into an FKS) gives  almost identical results to using an FKS directly.

\vspace{0.1in}

 \noindent As might be expected, the MMPDE approach with a high-order solver is best of all in terms of the approximation and results in an accurate piecewise linear approximation with errors of ${\cal O}(10^{-10})$ for the best approximations. This shows us that we can train a system get the expressivity we want with a piecewise linear approximation equivalent to a \ReLU \,NN, but the accuracy comes at a high computational cost.

\section{Generalisations} 

We have so far confined the theory, and experiments, to the case of the shallow \ReLU \,NN. This is done to allow a direct theoretical and numerical comparison with the linear spline FKS. We now make some preliminary investigations of extensions of these results to more general machine learning examples.

\subsection{More general activation functions}

\noindent 
In applications activation functions other than \ReLU~ are often used, such as leaky \ReLU, \ReLU-cubed, $\tanh$ and  sigmoid. As a  first calculation we consider the standard training, using Adam, of a shallow NN with $\tanh$ activation on the same target functions $u_i(x)$ as before, with the  $L^2_2$ loss. In Table \ref{tab:c2} we present the $L_2^2$ error when training this network with $N$ knots  and  $P$  parameters. The results for the smooth functions $u_1,u_2$ and mildly singular function $u_3$ are rather better than those for the  shallow \ReLU \,NN without preconditioning and this problem appears to be better conditioned than the \ReLU approximation in this case.   However we see  poor performance on the rapidly varying test function $u_4$, and the highly oscillatory test function $u_5$. If we consider the optimally trained ReLU NN (with the pre-conditioned two-level approach), for example when $N = 64$, as seen in the results presented in Section 8, we see errors of $10^{-10}$ or smaller in all of the examples. These errors are consistently much smaller than those in Table \ref{tab:c2}. 

\begin{table}[htb!] 
\centering
\begin{tabular}{|c|c|c|c|c|c|c|}  \hline
$N$ & $u_1$ & $u_2$ & $u_3$ & $u_4$ & $u_5$ & $P$ \\
\hline
$8$ & $4.6 \times 10^{-9}$ & $1.70 \times 10^{-6}$ & $3.55 \times 10^{-4}$ & $1.08 \times 10^{-3}$ & $1.77 \times 10^{-2}$ & $25$ \\ \hline
$16$ & $1.56 \times 10^{-9}$ & $1.20 \times 10^{-8}$ & $1.62 \times 10^{-7}$ & $8.01 \times 10^{-4}$ & $1.53 \times 10^{-2}$ & $49$ \\ \hline
$32$ & $1.21 \times 10^{-8}$ & $1.41 \times 10^{-8}$ & $1.50 \times 10^{-7}$ & $6.17 \times 10^{-4}$ & $1.64 \times 10^{-2}$ & $97$ \\ \hline
$64$ & $5.25 \times 10^{-9}$ & $8.88 \times 10^{-9}$ & $2.15 \times 10^{-7}$ & $5.54 \times 10^{-4}$ & $1.77 \times 10^{-2}$ & $193$ \\ \hline 
$128$ & $2.21 \times 10^{-8}$ & $8.00 \times 10^{-9}$ & $2.15 \times 10^{-7}$ & $4.40 \times 10^{-4} $ & $1.90 \times 10^{-2}$ & $385$ \\
\hline
\end{tabular}
\caption{The $L^2_2$ error of training a shallow network with the $\tanh$ activation function, without preconditioning, to approximate the test functions $u_i(x)$.}
\label{tab:c2}
\end{table}

 \subsection{Deeper networks}
 
 \noindent In approximations deep networks  are usually considered instead of shallow ones.  A deep network using \ReLU~activation is a piecewise linear function of its input, however, unlike the shallow case, the link between the knots of a FKS and the breakpoints of the deep network approximation is very subtle \cite{DeVore_Hanin_Petrova_2021} and it is much harder to develop a direct analytical theory in this case. Instead we briefly present some preliminary numerical comparisons and draw comparisons with the earlier analysed case of shallow networks.  For such a comparison we compare a 2-layer \ReLU NN  of width $W = 5$ and 46 trainable parameters,  with an FKS with $16$ knots and 49 trainable parameters.  The 2-layer NN was trained on the functions $u_i(x)$ using Adam with the $L^2_2$ loss function and no preconditioning.  Results are presented in Table \ref{tab:c3} in which we show the variance of results obtained with a number of initial choices of the parameters. In Table 2  we see poor convergence, similar to that  observed earlier for the shallow \ReLU \,NN. In all cases the errors are very much larger than the comparable two-level trained FKS.  The behaviour of the NN is rather erratic  and is critically dependant on the initial values of the parameters, suffering from the same issues with ill-conditioning as were observed earlier for the shallow \ReLU\,networks. Clearly, preconditioning deep networks is important for effective optimisation, and this should be the subject of future research.  

\begin{table}[htb!]
\centering
\begin{tabular}{|l|c|c|c|c|c|}  \hline
 & $u_1$ & $u_2$ & $u_3$ & $u_4$ & $u_5$ \\ \hline
$L_2^2$ error & $4.2 \times 10^{-3}$ & $5.47 \times 10^{-2}$ & $3.45 \times 10^{-2}$ & $1.8 \times 10^{-1}$ & $1.85 \times 10^{-2}$ \\ \hline
Variance & $9.71 \times 10^{-5}$ & $1.42 \times 10^{-2}$ & $1.35 \times 10^{-2}$ & $5.40 \times 10^{-2}$ & $1.09 \times 10^{-4}$ \\ 
\hline
\end{tabular}
\caption{Results of training a two-layer network with 5 nodes on the test functions.}
\label{tab:c3}
\end{table}

\section{Conclusions}\label{sec:conclusion}

 The results of this paper have shown that it is  hard to train a shallow \ReLU \,NN to give a particularly accurate approximation 
for either smooth or singular target functions if a standard method is used. In contrast, using a two-level equidistribution based training approach, combined with preconditioning it is certainly possible to train a \ReLU \,NN formally equivalent FKS to give very accurate approximations on the same set of target functions. Hence the high level of expressivity theoretically possible for the \ReLU \,NN approximation is achieved in training by the FKS and the preconditioned \ReLU \,NN, but not, in practical training using standard methods, by the standard \ReLU \,NN. The reason for this seems to be two-fold. Firstly, it is necessary to have (in both the FKS and the \ReLU \,NN) control over the knot location using (for example) an equidistribution based loss function. Secondly, the process of training the weights of the FKS is much better conditioned than that of finding the scaling coefficients of the \ReLU~functions which have a global support. As we have seen, the combination of these two factors leads to poor \ReLU \,NN approximations. Both issues can be overcome when training a \ReLU \,NN by using an equidistribution based loss function and then preconditioning the \ReLU problem to give it the same structure as an FKS. By doing this we can then reliably realise the full expressivity of the \ReLU \,NN approximation. 
These results have implications in the use of \ReLU \,NN to approximate functions in the context of a PINN and other aspects of (scientific) machine learning.

\vspace{0.1in}

\noindent In this paper we have deliberately mainly confined ourselves to piecewise linear function approximations in one spatial dimension by shallow networks. This is because we can then make a direct comparison between a shallow \ReLU \,NN and an FKS, and can develop both effective loss functions, and preconditioning strategies, which ensure that the optimisation approaches work well and delivers a highly accurate approximation. 

\vspace{0.1in}

\noindent Future work developing the results of this paper include (i) developing preconditioning strategies for deep univariate \ReLU \,NN networks, (ii) extending the results to higher dimensions, (iii) using the results to understand the convergence of PINN approximations.


\bibliographystyle{siam}      
\bibliography{references}

\appendix

\section{Example on optimal knots for the interpolating FKS}\label{ex:optimal}

\vspace{0.1in}

\noindent We consider the optimal knots for the interpolating FKS for a specific target function. 

\vspace{0.1in}

\begin{ex} 
\noindent As an   example, we consider  the target function 
$u(x) = x^{\alpha}$ for some $ \alpha \in(0, 1),$
which has a derivative singularity at $x = 0$.
We have
$$u''(x)^2= \alpha^2 (1 - \alpha)^2 x^{2\alpha - 4}.$$
It follows that the equidistributed set of mesh points satisfies the ODE
$$\frac{dx}{d\xi}  \alpha^{2/5} (1 - \alpha)^{2/5} x^{(2\alpha - 4)/5} = D, \quad x(0) = 0, \quad x(1) = 1.$$
Integrating with respect to $\xi$ 
yields
$$\frac{5}{2\alpha + 1} \alpha^{2/5} (1 - \alpha)^{2/5} (x(\xi))^{(2\alpha + 1)/5} = D\xi ,$$
implying that
$$x(\xi) =\left(\frac{2\alpha +1}{5} D \xi\right)^{5/(2\alpha+1)} \left( \alpha (1-\alpha) \right)^{-2/(2\alpha +1)}$$
As $x(1) = 1$, we   have
\begin{equation*}
    x(\xi) =  \xi^{5/(2\alpha + 1)}, \quad k_i = (i/(N-1))^{5/(2\alpha + 1)}, \quad D^5 = \alpha^2 (1 - \alpha)^2 (5/(2\alpha + 1))^{5}.
    \label{coct25}
\end{equation*}
Observe that $D = {\mathcal O}(1)$ so that the discretisation $L_I$ of $\mathcal L^2_2$ with optimal knots for an interpolating FKS is ${\cal O}(N^{-4})$.  For comparison, note that the discretisation $L_I$ of $\mathcal L^2_2$ on a uniform set of $N$ knot points of spacing $1/(N-1)$ satisfies
\begin{equation}
    L_I(\mathbf{\theta}) \approx \sum_{i=0}^{N-2}\frac{(u''(x(\xi_{i+1/2})))^2}{120 (N-1)^5} = \sum_{i=0}^{N-2}\frac{\alpha^2 (1 - \alpha)^2 \xi_{i+1/2}^{2\alpha - 4}}{120 (N-1)^5} 
    \label{coct23}
\end{equation}
for an appropriate choice of $\xi_{i+1/2}$. As \eqref{coct23}  is dominated by the contribution in the first cell with $\xi_{1/2}$ of order ${\mathcal O}(1/N)$, this implies that 
\begin{equation}
L_I(\mathbf{\theta})={\mathcal O}(N^{4 - 2\alpha} N^{-5}) = {\mathcal O}(N^{-1 - 2\alpha}).
\label{cfeb6a}
\end{equation}
In particular, the approximation error of the optimal knot choice is smaller than the one for the uniform set of knot points for any $\alpha\in(0,1)$. For $\alpha= 2/3$, for instance, \eqref{coct23} reduces to $k_i = (i/(N-1))^{15/7}$ and by the definition of the weights $w_i=u(k_i)$ for the interpolating FKS we obtain $w_i = (i/(N-1))^{10/7}$. 
Observe that $k_i$ and $w_i$ are bounded, and that $k_{i} - k_{i-1}$ rapidly decreases from $i=1$ to $i=0$.
The associated coefficients $c_i$ of the ReLU NN are shown for $N = 64$  in Figure \ref{fig:nnee70}. 
 \begin{figure}[htbp!]
    \centering
    \includegraphics[width=0.7\textwidth]{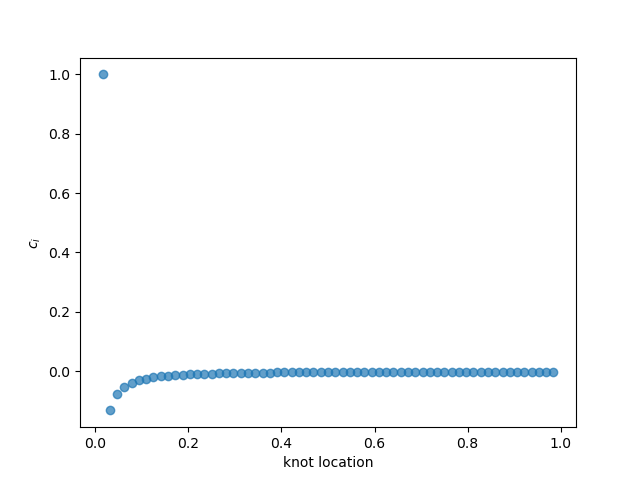}
    \caption{The unbalanced scaling coefficients $c_i$ of the ReLU NN approximation equivalent to the Interpolating FKS for the target function $x^{2/3}$.}
    \label{fig:nnee70}
\end{figure}
As expected, $c_0 > 0 > c_1$ and the values of $c_i$ are large for small $i$. The coefficients $c_i$ of the ReLU NN expression of the approximation of $u = x^{2/3}$ show rapid changes, and this adds to the ill-conditioning problems for finding the values of  $c_i$ identified earlier.  
\end{ex}

\end{document}